\documentclass[10pt,twocolumn,letterpaper]{article}

\usepackage{wacv}
\usepackage{times}
\usepackage{graphicx}
\usepackage{amsmath}
\usepackage{amssymb}
\usepackage{subfigure}
\usepackage{tablefootnote}
\usepackage{multirow}
\usepackage{url}
\usepackage{color}
\usepackage{adjustbox,lipsum}
\usepackage[vlined,boxed,commentsnumbered]{algorithm2e}

\newcommand{\dX}{\mathcal{X}}
\newcommand{\data}{\mathcal{D}}
\newcommand{\seg}{\mathcal{S}}

\newcommand{\loss}{\ell}
\newcommand{\Loss}{\mathcal{L}}
\newcommand{\setto}{\leftarrow}

\newcommand{\set}[1]{\left\{#1\right\}}
\newcommand{\expect}{\mathbb{E}}
\newcommand{\follows}{\sim}
\newcommand{\norm}[1]{\left\|{#1}\right\|}

\newcommand{\lftrt}{\leftrightarrow}

\newcommand{\tb}[1]{\textbf{#1}}
\newcommand{\rb}[1]{\rotatebox{90}{#1}}

\wacvfinalcopy % *** Uncomment this line for the final submission

\def\wacvPaperID{233} % *** Enter the wacv Paper ID here

\ifwacvfinal\fi
\begin{document}

%%%%%%%%% TITLE
\title{Sem-GAN: Semantically-Consistent Image-to-Image Translation}

% Authors at the same institution
\author{Anoop Cherian \hspace{2cm} Alan Sullivan \\
Mitsubishi Electric Research Labs (MERL), Cambridge, MA\\
{\tt\small \{cherian, sullivan\}@merl.com}
}

\maketitle
\ifwacvfinal\thispagestyle{empty}\fi

%%%%%%%%% ABSTRACT
\begin{abstract}
Unpaired image-to-image translation is the problem of mapping an image in the source domain to one in the target domain, without requiring corresponding image pairs. To ensure the translated images are realistically plausible, recent works, such as Cycle-GAN, demands this mapping to be invertible. While, this requirement demonstrates promising results when the domains are unimodal, its performance is unpredictable in a multi-modal scenario such as in an image segmentation task. This is because, invertibility does not necessarily enforce semantic correctness. To this end, we present a semantically-consistent GAN framework, dubbed~\emph{Sem-GAN}, in which the semantics are defined by the class identities of image segments in the source domain as produced by a semantic segmentation algorithm. Our proposed framework includes  consistency constraints on the translation task that, together with the GAN loss and the cycle-constraints, enforces that the images when translated will inherit the appearances of the target domain, while (approximately) maintaining their identities from the source domain.  We present experiments on several image-to-image translation tasks and demonstrate that Sem-GAN improves the quality of the translated images significantly, sometimes by more than 20\% on the FCN score.  Further, we show that semantic segmentation models, trained with synthetic images translated via Sem-GAN, leads to significantly better segmentation results than other variants.
%\keywords{generative adversarial networks, image-to-image translation}
\end{abstract}
\section{Introduction}
\label{sec:intro}
The recent advancements in several fundamental computer vision tasks~\cite{he2016deep,carreira2017quo,chen2016deeplab} are unequivocally associated with the availability of huge annotated datasets for training deep architectures of increasing sophistication~\cite{deng2009imagenet,heilbron2015activitynet,lin2014microsoft}. However, collecting or annotating such datasets are often challenging or expensive. An alternative, which is cheap and manageable, is to resort to computer gaming software~\cite{de2017procedural,richter2016playing,ros2016synthia} to render realistic virtual worlds; such software could supply unlimited amounts of training data and could also simulate real world scenarios that may be otherwise difficult to observe. Unfortunately, using data from a synthetic domain often introduces biases in the learned model, resulting in a domain shift that might hurt the performance of a downstream task~\cite{patel2015visual,wang2018deep}. 

\begin{figure}[htbp]
\centering
\subfigure[Real$\to$Synthetic]{\includegraphics[width=2.6cm]{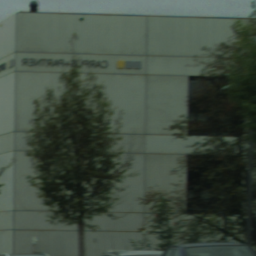}}
\subfigure[Cycle GAN]{\includegraphics[width=2.6cm]{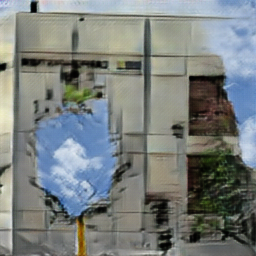}}
\subfigure[Sem-GAN (Ours)]{\includegraphics[width=2.6cm]{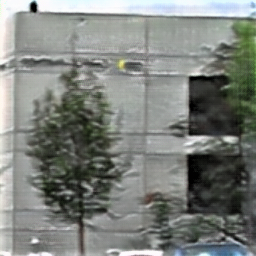}}
\caption{Translation of an image from the Cityscapes dataset (leftmost) to a synthetic domain. The state-of-the-art model Cycle-GAN~\cite{zhu2017unpaired} incorrectly maps 'trees' to `sky'. Using Sem-GAN, that enforces semantic consistency in the translation process, results in more realistic translations. }
\label{fig:comparison_figure}
\end{figure}

A standard way to account for domain shift is to adapt the synthetic images so that their statistics match that of the real domain. This is the classical domain adaptation problem~\cite{bousmalis2017unsupervised,ganin2015unsupervised,fernando2013unsupervised}, commonly called image-to-image translation when done on image pixels~\cite{rosales2003unsupervised,zhu2017unpaired,shrivastava2017learning}. Most such translation algorithms require corresponding image pairs from the two domains~\cite{isola2017image,eigen2015predicting,johnson2016perceptual}. However, thanks to generative adversarial networks (GAN), recent years have seen breakthroughs in unsupervised translations that do not require paired examples, instead need only sets of examples from the two domains, which are much easier to obtain~\cite{shrivastava2017learning,liu2016coupled,liu2017unsupervised,zhu2017unpaired}. However, the lack of correspondences results in a harder problem to solve, as one needs to estimate the image distributions and the adaptation function from the two sets -- an ill-posed problem, since an infinite number of marginal distributions are possible that could have generated the finite examples in each of these sets. 
% ; thus translations could be achieved by first embedding the inputs to this latent space, followed by conditioning the GAN generators on their embeddings

To ameliorate this intractability, recent methods make assumptions on the problem domains or the mapping function. For example, in Liu et al.~\cite{liu2016coupled,liu2017unsupervised}, the two domains are assumed to share a common latent space. In Cycle GAN~\cite{zhu2017unpaired}, the mapping is assumed invertible, i.e., a translated image when mapped back must be the same as the input image. Learning such a mapping could avoid some well-known pitfalls in GAN training such as mode collapse, and could allow learning a bijective mapping between the domains.

When dealing with real-world tasks, bijective mappings might not be sufficient to generate meaningful translations. For example, consider the real-to-synthetic translation task depicted in Figure~\ref{fig:comparison_figure}. Here, the Cycle-GAN is trained on real images from the Cityscapes dataset~\cite{cordts2016cityscapes} and synthetic road scene images are produced by the Mitsubishi Precision Simulator.\footnote{\url{https://www.mpcnet.co.jp/e/e_product/sim/index.html}} As is clear, Cycle-GAN has learned an incorrect mapping between the classes 'trees' and 'sky', resulting in implausible translations. Nevertheless, such a mapping is invertible as per the Cycle-GAN cost function. This problem happens because, in typical translation tasks the images in the two sets are assumed to be samples from the joint distribution of all their respective sub-classes (object segments), and the translation is a mapping between such joint distributions. Such a mapping (even with cycles) does not ensure that the marginals of the sub-classes (modes) are assigned correctly (e.g., sky$\to$sky). To this end, we try to look beyond cyclic dependencies, and incorporate semantic consistency as well in the translation process.

In this paper, we propose a novel GAN architecture for pixel-level domain adaptation, coined semantically-consistent GAN (Sem-GAN), that takes as input two unpaired sets, each set consisting of tuples of images and their semantic segment labels, and learns a domain mapping function by optimizing over the standard min-max generator-discriminator GAN objective~\cite{goodfellow2014generative}. However, differently from the standard GAN, we pass the generated images through a semantic segmentation network~\cite{long2015fully} (in addition to the discriminator); this network trained to segment images in the target domain. It is expected that if the translation is ideal, the objects being translated will inherit their appearance from the target domain while maintaining their identity from the source domain. For example, when a 'car' class in one domain is translated, a segmentation model trained to identify the 'car' class in the target domain should identify the translated object in the same class. We use the discrepancy between the ground-truth semantic classes and their predictions as error cues (via cross-entropy loss) for improving the generator via backpropagation. 

Given that semantic segmentation by itself is a difficult (and unsolved) computer vision problem, a natural question would be how useful it can be to include such an imperfect module in a GAN setup. Our experiments show that using a segmentation scheme that performs reasonably well (such as FCN~\cite{long2015fully}) is sufficient to ensure semantic consistency, leading to better translations. Further, we also propose to train the segmentation module jointly with the discriminator; as a result, its accuracy improves along with the generator-discriminator pair. A careful reader would have noticed, we are in fact solving a \emph{chicken-egg problem}: on the one hand, we use GAN to improve semantic segmentation, while on the other hand, we are using segmentation to improve image-to-image translation. To clarify, we do not assume an accurate segmentation model, instead some model that performs reasonably well will suffice, which could be obtained via training a semantic segmentation model in a supervised setup using limited data; for example, we use about 1K annotated images in our experiments on the Cityscapes dataset. Our goal is to use this model to improve domain adaptation, so that we can adapt a large number of synthetic images to the target domain, to train a better segmentation model on the target domain. 

The use of the segmentation models could help us even better. As alluded to above, the main challenge in standard image translation models is the inability of the network to find the correct mode mapping. We explore this facet of our framework by introducing semantic dropouts by stochastically blanking out semantic classes from the inputs so that the network can learn to map specific classes independently. We present experiments on a variety of image-to-image translation tasks and show that our scheme outperforms those using cycle-GAN significantly.  %We envisage that once the GAN setup is guided to generate semantically-consistent translations, it could then be applied for situations when ground truth labels are absent from the target class -- a problem tackled in several other concurrent works~\cite{hoffman2017cycada,shrivastava2017learning,ma2018gan,gan2017triangle}. 
% (for example, guiding them to learn the correct mapping between `sky' and 'trees')

Before moving on, we summarize below our main contributions:
\begin{itemize}
\item We introduce a novel feedback to the generator in GANs using predictions from a semantic segmentation model. %Our scheme forces the generator to produce realistic object appearances that maintain their class identities.

\item We propose a GAN architecture that includes a segmentation module and the whole framework trained in an end-to-end manner.

\item We introduce semantic dropout for improving our consistency loss.

\item We present experiments on several image-to-image translation tasks, demonstrating state-of-the-art results (sometimes by more than 20\% in FCN score against Cycle-GAN) Further, we provide experiments using the proposed translation for training semantic segmentation models using large synthetic datasets, and show that our translations lead to significantly better segmentation models than state-of-the-art models (by 4-6\% in mean IoU score).
\end{itemize}

\section{Related Work}
\label{sec:related_work}
% ; one learning to generate data from the true data distribution, while the other learning to discriminate the generated data against the true data; the generator and the discriminator reach a saddle point in the optimization landscape; i.e.,
GANs~\cite{goodfellow2014generative,salimans2016improved,radford2015unsupervised,arjovsky2017wasserstein} allow learning complex data distributions automatically by pitting two CNNs (a generator and a discriminator) against each other using an adversarial loss. The optimum for this non-convex min-max game is when the generator produces data which the discriminator cannot distinguish from real data. This key idea has led to major advancements in applications requiring data synthesis such as: representation learning~\cite{radford2015unsupervised}, image generation~\cite{wang2017high,wang2016generative,radford2015unsupervised}, text-to-image synthesis~\cite{reed2016generative}, inpainting~\cite{pathak2016context}, super-resolution~\cite{ledig2016photo}, face-synthesis~\cite{liu2017face}, style-transfer~\cite{johnson2016perceptual,wang2016generative,rosales2003unsupervised}, and image editing~\cite{zhu2016generative}. %, and video generation~\cite{xiong2017learning,vondrick2016generating,mathieu2015deep}. 
% image colorization~\cite{zhang2016colorful,he2017neural}, 3D shape synthesis~\cite{tung2017adversarial,wu2016learning},  texture synthesis~\cite{ulyanov2016texture},

The basic GAN~\cite{goodfellow2014generative,radford2015unsupervised} framework is extended in Liu and Tuzel~\cite{liu2016coupled} to model the joint distribution of paired images from two distinct domains by coupling two GANs by sharing their weights. This scheme is extended in VAE-GAN~\cite{liu2017unsupervised} for image-to-image translation using auto-encoders to embed images to a shared space, on which the generators are conditioned. SimGAN~\cite{shrivastava2017learning} replaces the noise input in traditional GANs by synthetic images, and asking the generator to refine these images to look as real as possible. Similar to ours, SimGAN uses an FCN for  translation consistency; however, this FCN is used to preserve the holistic structure of the images and not to preserve the class identities. In Cycle-GAN~\cite{zhu2017unpaired} and dual-GAN~\cite{yi2017dualgan}, the translations are required to be bijective. However, as noted earlier, such bijective mappings need not preserve semantic consistency.
%  real domain through a self-regularization loss,
% However, it is unclear if this shared space can preserve the semantics of its inputs without additional constraints.
%The scheme demonstrates strong empirical performance on unimodal translation tasks, however it is theoretically unclear (without additional assumptions and constraints) how the method would perform in a multi-modal translation setup.
%The recent Pix2Pix framework~\cite{isola2017image} learns semantically-consistent translations, however requires the source and target data to be paired.
Other forms of consistencies have been explored in recent works. In Sangkloy et al.~\cite{sangkloy2017scribbler}, consistency between sketch boundaries is presented, perceptual consistency is enforced in DeepSim~\cite{dosovitskiy2016generating}, while geometric consistency is explored in Xu et al.~\cite{xu2018generating}. Self-similarity is used in Deng et al.~\cite{deng2017image}, and feature-level consistency is assumed in X-GAN~\cite{royer2017xgan}. In Luo et al.~\cite{luo2017label}, domain alignment is used. 

There are several works concurrent to ours that explore other ways for semantic consistency. In PixelDA~\cite{bousmalis2017unsupervised}, DTN~\cite{taigman2016unsupervised}, and CyCADA~\cite{hoffman2017cycada}, the translation task is required to be invariant under a pre-defined criteria -- such as a classifier performance. However, their tasks are different from ours and do not assume the availability of target segmenters like we do. Semantic consistency using attention is presented in DA-GAN~\cite{ma2018gan}. In Sankaranarayan et al.~\cite{sankaranarayanan2017unsupervised}, a neighborhood-preserving feature embedding is introduced. Similarly Li et al.~\cite{li2018semantic} uses a semantic-aware discriminator to preserve the high-level appearances. In triangle GAN~\cite{gan2017triangle}, a semi-supervised approach is presented for paired examples. In bidirectional-GAN~\cite{russo2017source}, the source class labels are used for semantic consistency. In one-sided GAN~\cite{benaim2017one}, the cycle-loss is replaced by neighborhood constraints. In comparison to these works, we tackle a different problem in which we assume to have access to (approximate) segmentation networks that can extract the source and target class labels.% Further, we assume the source and target domains are unpaired against each other.
%(albeit we may not have very large amounts of annotated target data, but just sufficient to train a reasonable segmentation network)
%In symmetric bidirectional-GAN~\cite{russo2017source}, the source class labels are used to enforce semantic consistency; i.e., the translated source when translated back should retain their original class labels -- similar to cycle-consistency.

To summarize, while semantic consistency has been explored from multiple facets in prior works, we are unaware of any work that explores segmentation consistency in the way we propose. We believe ours is the first work that leverages advances in the segmentation arena into the GAN framework for the image-to-image translation problem.

\section{Proposed Method}
\label{sec:proposed_method}
In this section, we present our Sem-GAN framework. To set the stage, we first review in the following sections, important previous work on GAN and Cycle-GAN on which our scheme is based. 

\subsection{Problem Setup}
\label{sec:setup}
Let $\data_A,\data_B$ be two image domains and let $\dX_A,\dX_B$ be sets of samples (images) from each domain respectively. Further, let $x_A\in\dX_A$ and $x_B\in\dX_B$ denote data samples. We assume $\dX_A$ and $\dX_B$ are unpaired, however, the two domains share $K$ semantic segment classes (with plausibly varied appearances). Assuming $\seg$ is the space of all segmentation masks with $K$ classes, let $\seg_A:\data_A\to \seg$ and $\seg_B:\data_B\to\seg$ be two functions mapping each pixel in each input image to their respective class label in the segmentation mask. In case, if we have access to the ground-truth masks, then we use $g_A$ and $g_B$ to denote these masks for inputs $x_A$ and $x_B$ respectively. Idealy, $g_A=\seg_A(x_A)$ and $g_B=\seg_B(x_B)$. In this case, $(x_A, g_A)$ and $(x_B, g_B)$ form an image-ground-truth pair from each domain (however note that the pairs $(x_A,g_A)$ and $(x_B, g_B)$ remain unpaired). 
%As is clear by now, 
%Following the recent success in unsuperivsed image-to-image translation~\cite{shrivastava2017learning,zhu2017unpaired,liu2017unsupervised}, we use a GAN framework as our backbone for learning the inter-domain mapping. 

\subsection{Generative Adversarial Networks}
A standard GAN~\cite{goodfellow2014generative} consists of two convolutional neural networks (CNN), termed a \emph{generator} and a \emph{discriminator}; the former takes random noise as input to produce an image, while the latter identifies if its input is a true or a generated image. The parameters of the generator and discriminator CNNs are optimized against an adversarial loss in a min-max game~\cite{arjovsky2017wasserstein,goodfellow2014generative,radford2015unsupervised}. 

\begin{figure}[ht]
\centering
\subfigure[GAN-Loss]{\label{fig:gan-loss}\includegraphics[width=5.5cm,trim={4.5cm 2.5cm 4cm 3cm}, clip]{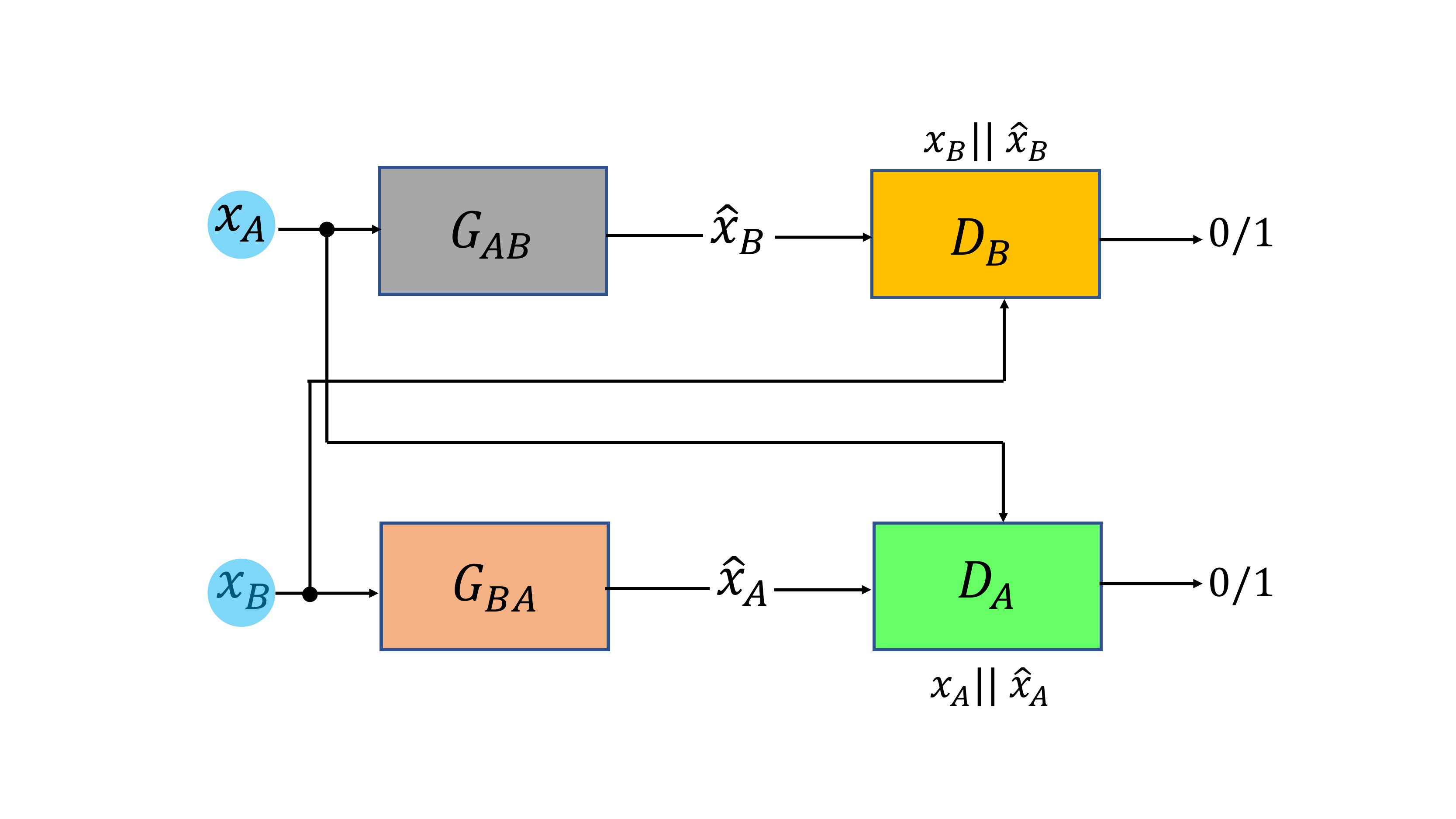}}
\subfigure[Cycle-Loss]{\label{fig:cycle-loss}\includegraphics[width=5.5cm,trim={3.0cm 2.5cm 1.5cm 4cm},clip]{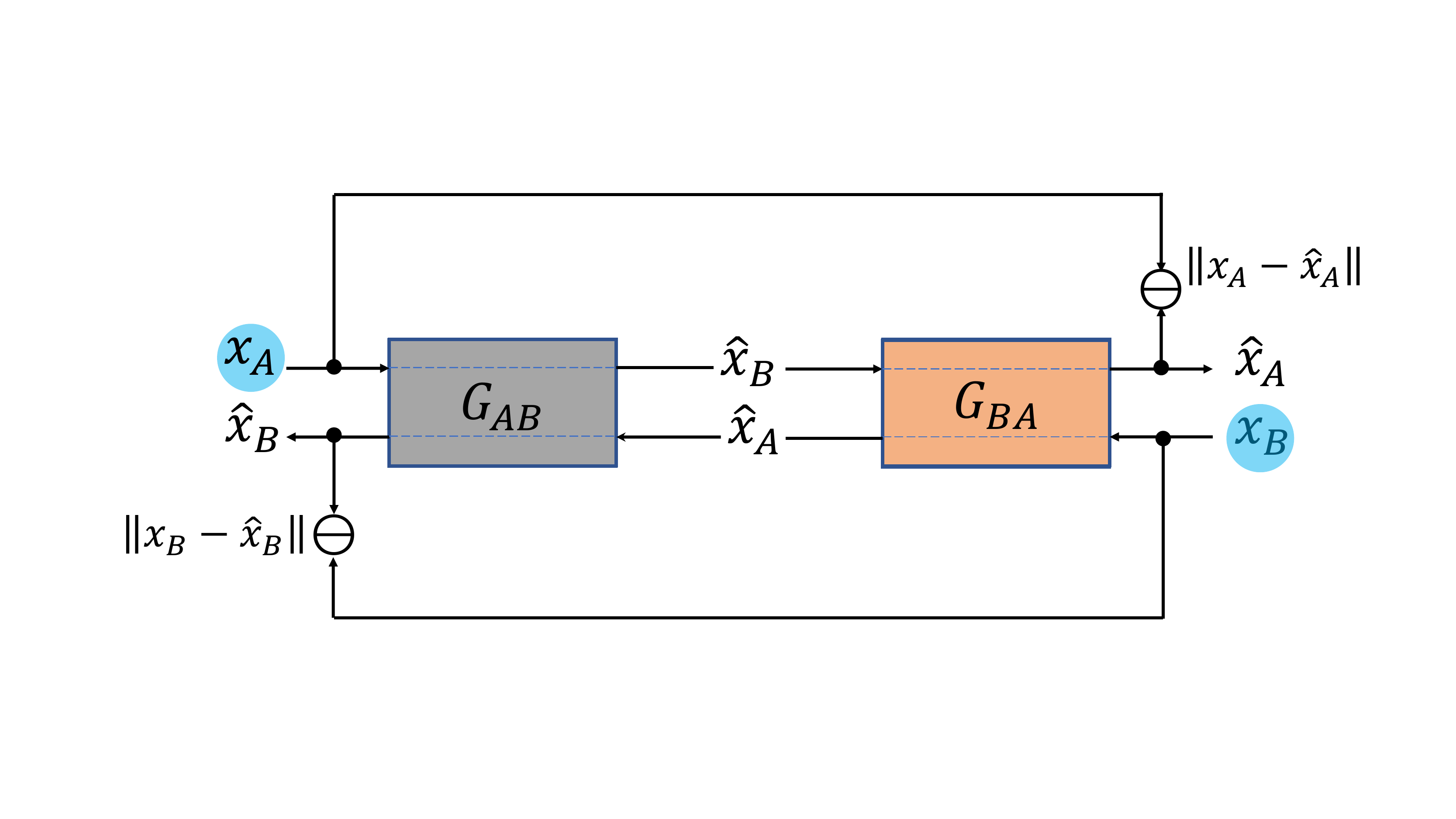}}
\caption{Illustrations of the objectives. The inputs are highlighted in `blue' circles.}
\label{fig:losses}
\end{figure}

Extending this idea to an image-to-image translation setting, we define two generators $G_{AB}$ and $G_{BA}$ defined as $G_{AB}:\dX_A\to \data_B$ and $G_{BA}:\dX_B\to\data_A$ and two binary discriminators $D_A$ and $D_B$, where $D_A(x)=1, \forall x\in\dX_A$ and $D_A(x)=0,\forall x\in G_{BA}(\dX_B)$. Similarly, $D_B(x)=1,\forall x\in\dX_B$ and $D_B(x)=0, \forall x\in G_{AB}(\dX_A)$. Here, with a slight abuse of notation, we assume $G(\dX)$ is the set of fake images produced by a generator $G$ from domain $\dX$. To learn the parameters of the generators and the discriminators, we define the following adversarial losses using binary cross-entropy:

\begin{align}
 \min_{G_{AB}} \max_{D_B}\quad \loss^{AB}_{GAN} &:= \expect_{x_B\!\follows\!\data_B}\!\log(D_B(x_B)) + \notag\\&\expect_{x_A \follows \data_A} \log(1-D_B(G_{AB}(x_A))),\label{eq:gan1}\\
\min_{G_{BA}} \max_{D_A}\quad \loss^{BA}_{GAN} &:= \expect_{x_A\follows \data_A}\log(D_A(x_A)) +\notag\\&\expect_{x_B\follows \data_B}\log(1-D_A(G_{BA}(x_B))).
\label{eq:gan2}
\end{align}
The GAN architecture for these objectives is graphically illustrated in Figure~\ref{fig:gan-loss}. While,~\eqref{eq:gan1} and~\eqref{eq:gan2} represent a non-convex game whose optimum parameters correspond to saddle points ( and thus typically difficult to optimize) it is often seen that with suitable heuristics~\cite{goodfellow2014generative,radford2015unsupervised,salimans2016improved} and careful choices for the loss ~\cite{arjovsky2017wasserstein,mao2017least}, the problem converges to practically useful solutions.  

\subsection{Cycle-Consistent GAN}
An important problem that one often encounters when training GANs is that of mode collapse which happens when the generators $G$ learn to produce a few samples from the true data domain without completely spanning it. Given that the discriminator loss does not enforce diversity in its inputs; the optimization may converge to such local solutions. Among several workarounds proposed to tackle this problem~\cite{arjovsky2017wasserstein,radford2015unsupervised,salimans2016improved}, one that has been very promising in the image-translation setting is that of Cycle-GAN~\cite{zhu2017unpaired} that puts additional constraints to the GAN objective enforcing diversity implicitly. Specifically, the Cycle-GAN loss asks for the translated data to be re-translated back to their original inputs. Mathematically, this loss can be written as:
\begin{align}
\loss_{cycle}:= &\expect_{x_A\follows\data_A}\norm{x_A-G_{BA}(G_{AB}(x_A))} +\notag\\&\quad\expect_{x_B\follows\data_B}\norm{x_B-G_{AB}(G_{BA}(x_B))},
\label{eq:cycle}
\end{align}
where $\|\ . \ \|$ is a suitable norm. Optimizing this requirement within a GAN formulation~(\ref{eq:gan1},~\ref{eq:gan2}) automatically demands that the generators $G_{AB}$ and $G_{BA}$ learn unique mappings such that they may be invertible to the original inputs; thereby elegantly avoiding the collapse of the data modes. The cyclic-constraints are depicted in Figure~\ref{fig:cycle-loss}.

While, the cyclic constraint has resulted in several compelling image-to-image translation results on domains that are often uni-modal (such as horse-zebra, day-night etc.), it is theoretically unclear how the scheme may perform in a multi-modal setup. This is because, when there are multiple modes in either domains, any bijective mapping between the modes will satisfy the invertibility constraint in~\eqref{eq:cycle}, such as in the example depicted in Figure~\ref{fig:comparison_figure}. Recently, this assignment problem has been looked at in the work of Gananti et al.~\cite{tau2017role}, hypothesizing that by constraining the space of possible translation functions (by controlling the capacity of the CNNs) may be allowing them to learn minimal-complexity mappings. While, this result is interesting, it may be practically difficult to achieve. Instead, we seek to constrain the possible mappings by guiding the generators to learn mappings that are semantically consistent with respect to a segmentation loss, viz. Sem-GAN.

\subsection{Semantically-Consistent GAN}
\label{sec:seggan}
%As alluded to in Section~\ref{sec:intro}, our main idea in Sem-GAN is to preserve the class identity of segments in the source domain, while adapting their appearances to the target domain. Such a translation might not be appropriate for some domains, for example, when translating `horses' to `zebras'~\cite{zhu2017unpaired}. However, in several other tasks, such as synthetic to real transformation when using simulated road-scene images to train a vehicle navigation system, maintaining the semantics is of utmost importance. 

In this section, we present our Sem-GAN framework. Re-using notation from Section~\ref{sec:setup}, we assume to have segmentation functions $S_A$ and $S_B$ that are trained to segment images from their respective domains. We do not assume that these segmenters are trained on $\dX_A$ and $\dX_B$ necessarily, but could be trained on external datasets that are similar to our domains. Using these segmenters, our semantic consistency constraint is written as:
\begin{align}
\loss_{seg} := &\expect_{x_A\in\data_A}\ \Loss(S_A(x_A), S_B(G_{AB}(x_A))) + \notag\\ &\quad\expect_{x_B\in\data_B}\ \Loss(S_B(x_B), S_A(G_{BA}(x_B))),
\label{eq:segloss}
\end{align}
where $\Loss(P,Q)$ is a suitable loss comparing two segmentation masks $P,Q\in\seg$, e.g., the cross-entropy loss used in FCN~\cite{long2015fully}. Specifically, in~\eqref{eq:segloss}, we enforce that the segmentation of $x_A$ by a function $S_A$ trained using images from domain $\data_A$ should be preserved by a segmentation function $S_B$ trained on images from $\data_B$ when applied to the translated image. When ground truth semantic labels $g_A$ and $g_B$ are available for either domains, we replace $S_A(x_A)$ by $g_A$ and $S_B(x_B)$ by $g_B$ in~\eqref{eq:segloss}. In this case, Eq.~\eqref{eq:segloss} can be written as:
\begin{align}
\loss^g_{seg} :=\ &\expect_{\small{(x_A, g_A)\in\data_A\times \seg}}\quad \Loss(g_A, S_B(G_{AB}(x_A)))\notag\\
			&+ \expect_{\small{(x_B, g_B)\in\data_B\times \seg}}\quad \Loss(g_B, S_A(G_{BA}(x_B))).
\label{eq:seglossg}
\end{align}

\begin{figure}[ht]
\centering
\subfigure[Seg-Loss w/o ground truth]{\label{fig:seg-loss-1}\includegraphics[width=6cm,trim={4.5cm 1.5cm 1.8cm 1.5cm},clip]{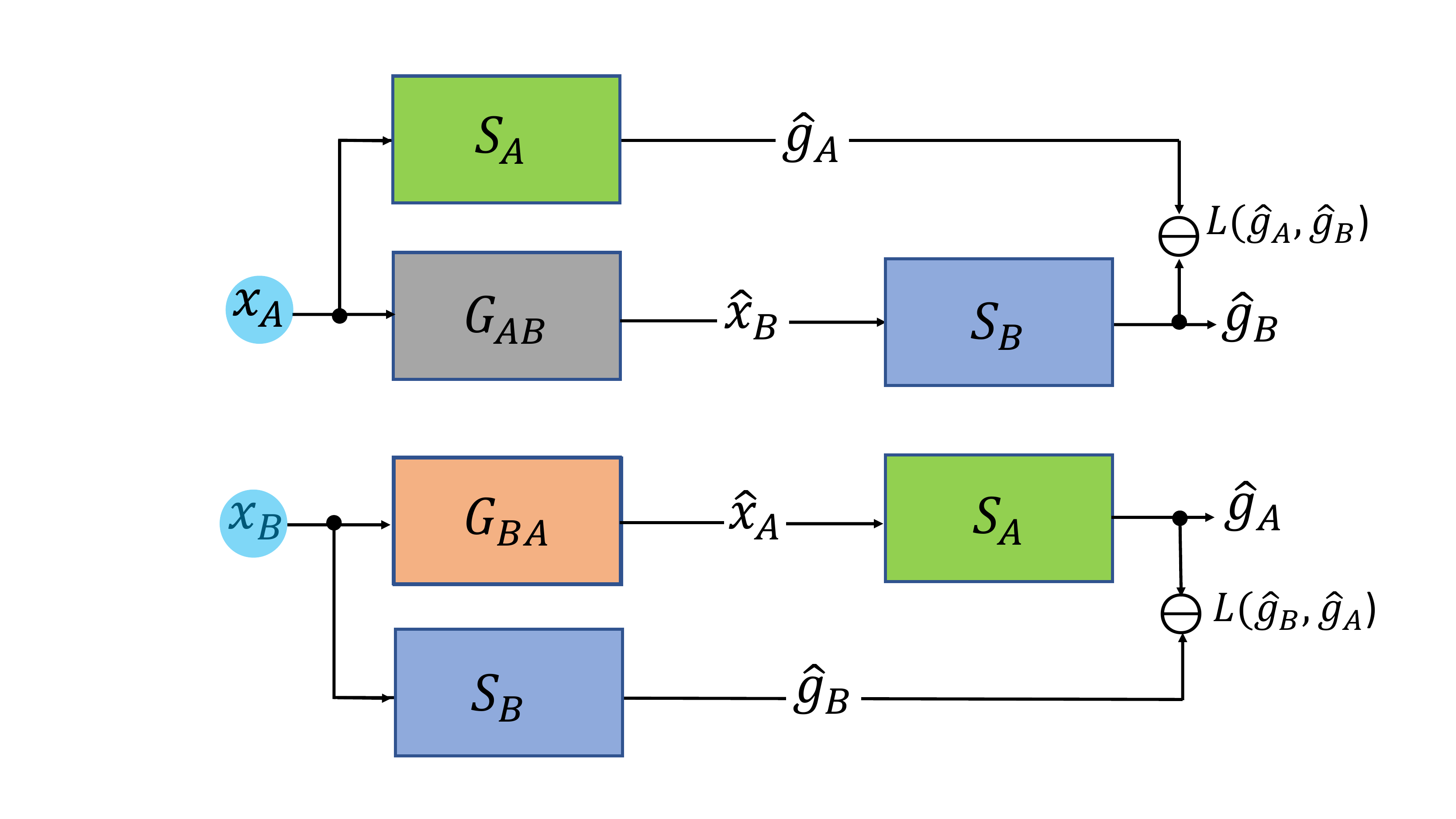}}
\subfigure[Seg-Loss w/ ground truth]{\label{fig:seg-loss-2}\includegraphics[width=6cm,trim={3.0cm 2.1cm 2cm 2cm},clip]{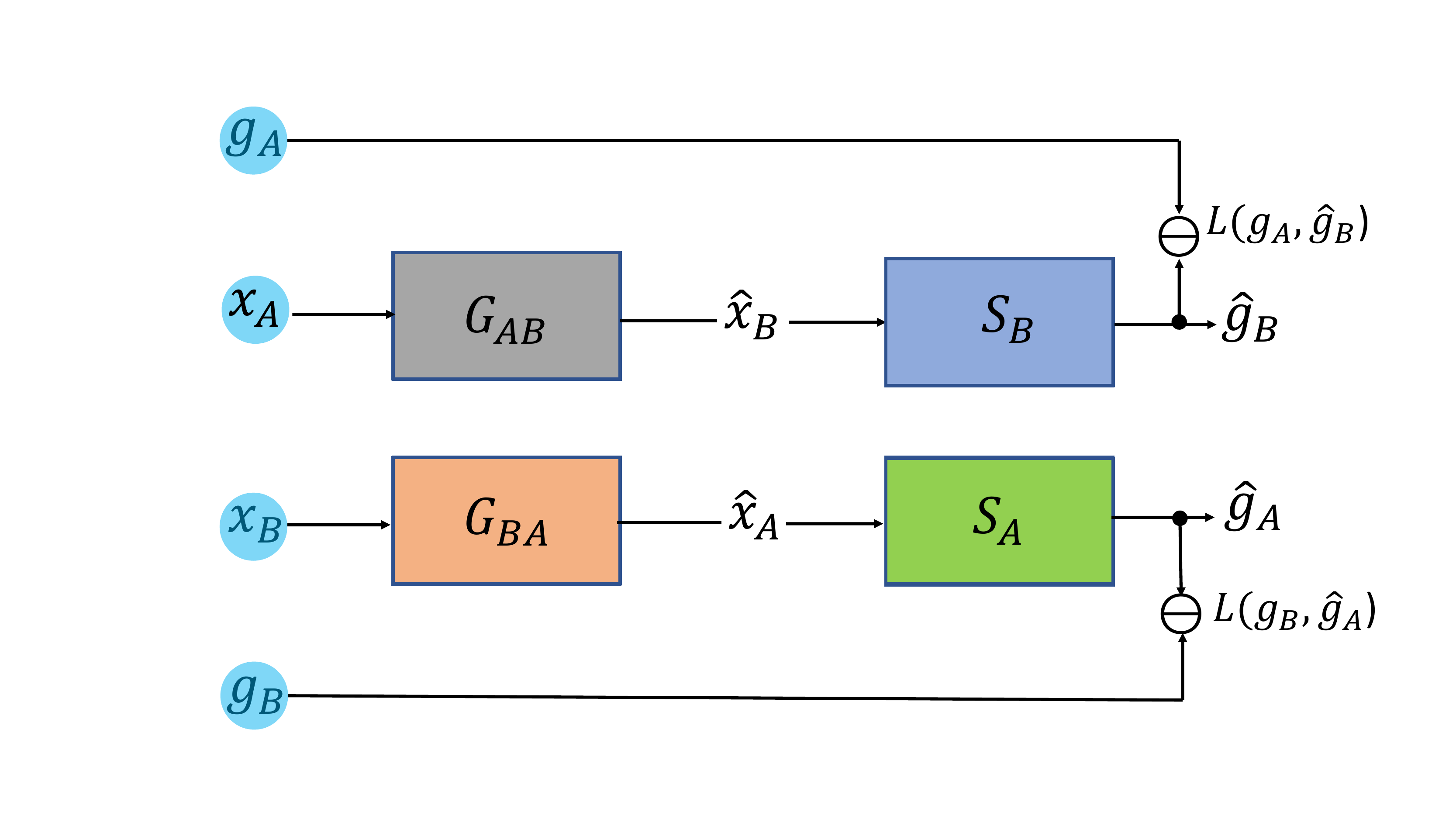}}
\caption{Two variants of our segmentation consistency model. Left: when the ground truth annotations are not used. Right: when they are used.}
\label{fig:seg-losses}
\end{figure}

Figure~\ref{fig:seg-loss-1} and~\ref{fig:seg-loss-2} illustrate these two variants of our Seg-GAN losses. For convenience in the depiction, we have introduced additional variables $\hat{g}$ to denote the outputs of the segmentation modules. 

\subsection{Overall Architecture}
\begin{figure}
\centering
\includegraphics[width=4cm,trim={1.8cm 4.4cm 8.4cm 0.5cm},clip]{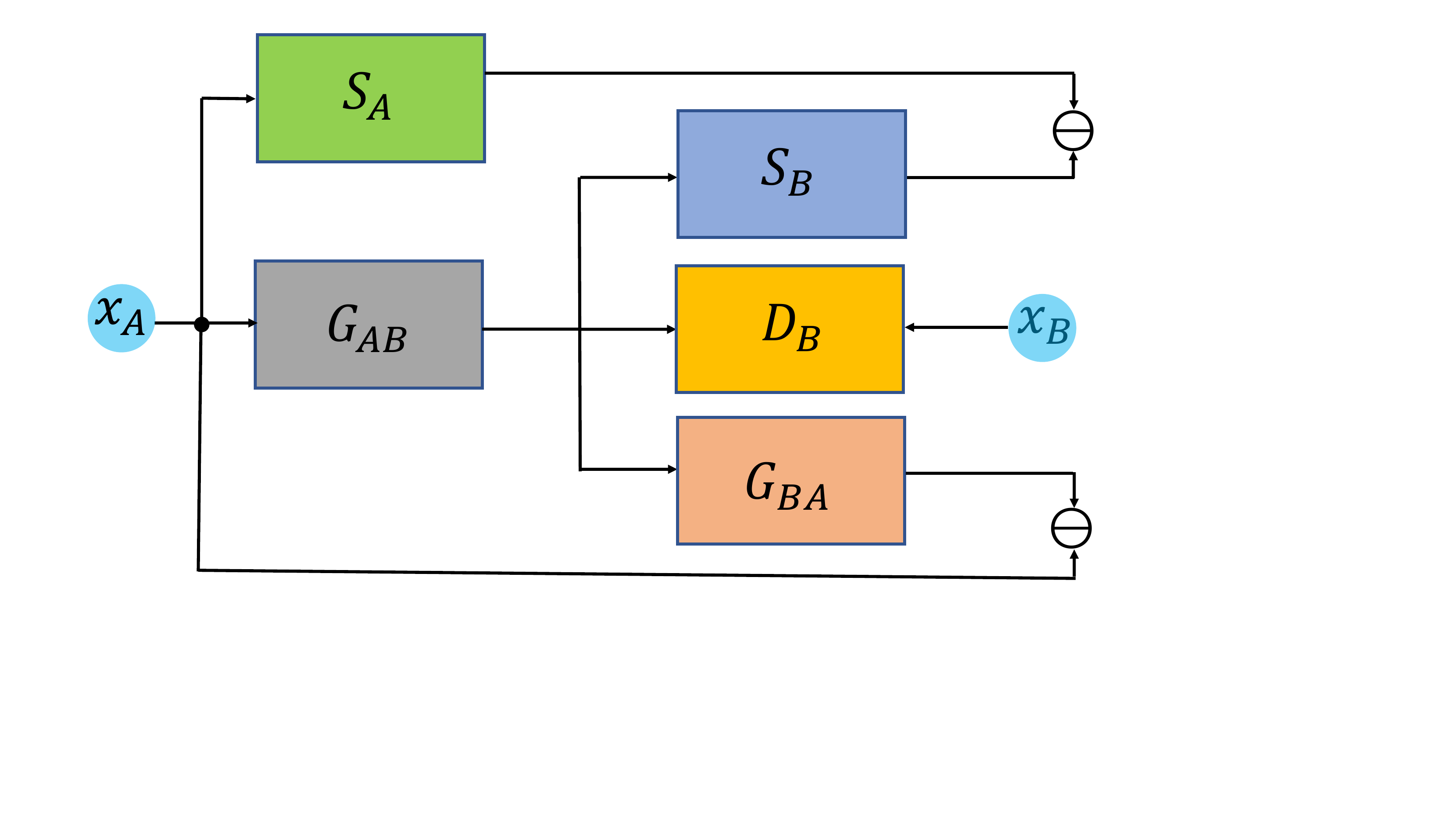}
\includegraphics[width=4cm,trim={1.8cm 4.4cm 8.4cm 0.5cm},clip]{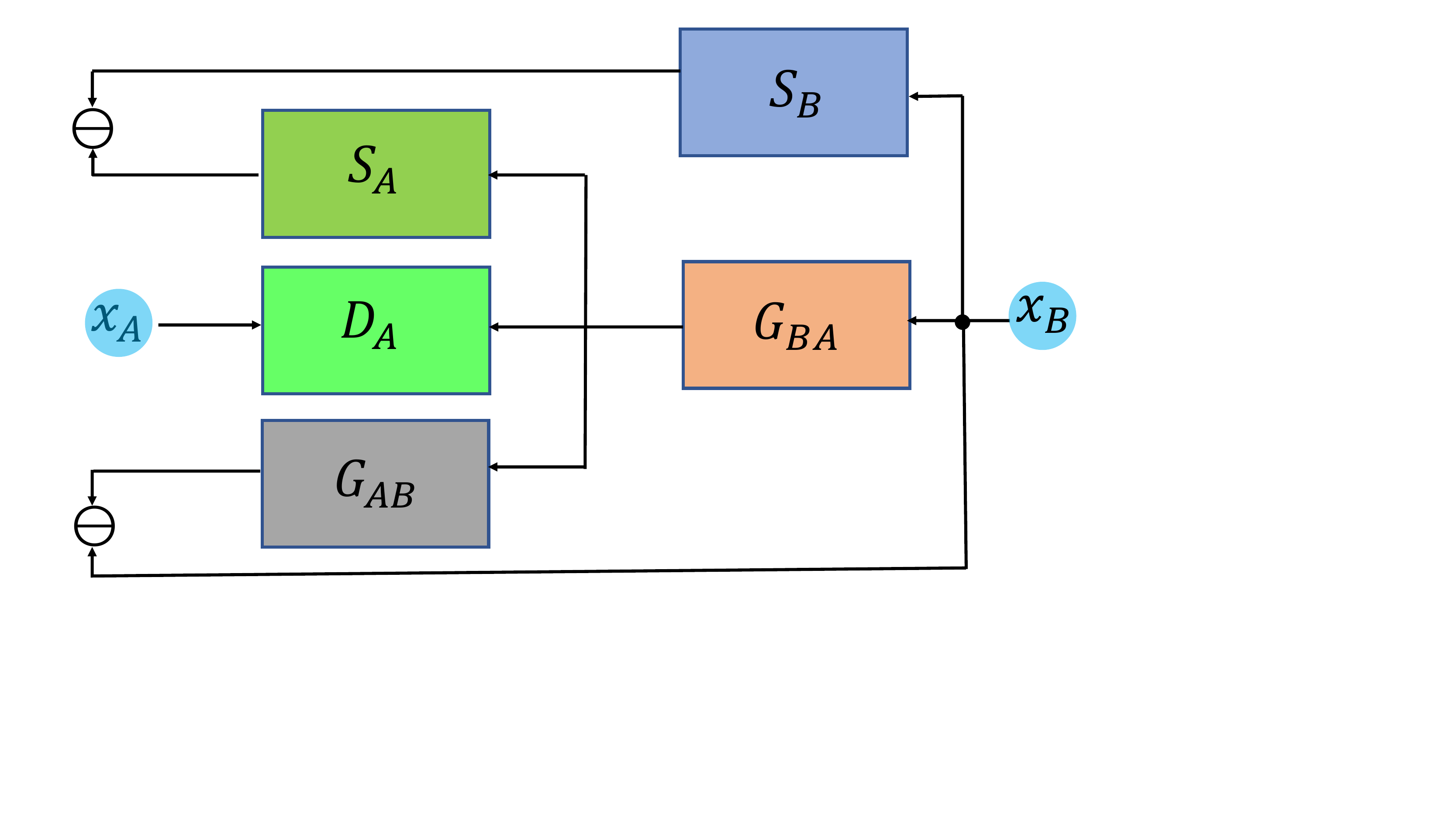}
\caption{Overall architecture. Note that some modules are repeated on the left and the right parts of our illustration to avoid cluttered cross-connections. Thus, blocks with the same color represent the same module.}
\label{fig:arch}
\end{figure}

In Figure~\ref{fig:arch}, we present our complete Sem-GAN architecture, combining the three losses. Notably, we include the two segmentation modules, $S_A,S_B$, as part of the setup; which are trained end-to-end alongside the generators and the discriminators. Using hyperparameters $\lambda_1$ and $\lambda_2$, our full loss is given by:
\begin{equation}
\loss := \loss^{AB}_{GAN} + \loss^{BA}_{GAN} + \lambda_1\loss_{cycle} + \lambda_2\loss_{seg}.
\label{eq:loss}
\end{equation}

With such an end-to-end architecture, that learns the segmenters and the discriminators together, there is a subtle but important risk. Note that, since $G_{AB}$ and $G_{BA}$ are learned alongside $S_A$ and $S_B$, it is often observed empirically that the segmenters will learn unrealistic appearances as valid ground truth classes when the generators are still in the learning phase. For example, suppose say in early epochs, the generator has not started translating valid 'car' images. Instead, it translates the appearance of a 'person' to a 'car' class (this is possible as we do not have paired data). Now, when back-propagating the error to update the parameters of the segmenter using the ground truth mask for the translated image, the segmenter may incorrectly learn to map the `person' appearance to 'car' class. This crucial issue may fail the semantic consistency.  To circumvent this problem, we propose to optimize the segmenters using their ground-truth labels alongside updating the discriminator parameters; i.e., we do not use the generator outputs to train the segmenters until they are accurate.
%  trim={<left> <lower> <right> <upper>}

\subsection{Semantic Dropout}
\label{sec:smdrop}
The availability of segmenters $S_A$ and $S_B$ allow for applying some modifications to our inputs such that the translation networks can be trained more effectively. Specifically, we could arbitrarily mask out object classes from both the input images. As a result, a generator could learn to map corresponding classes individually (mode-to-mode translation rather than translating the joint distribution of all labels together). Precisely, let $M^k$ denote a segment mask for class $k\in\set{1,2,\cdots, K}$, which has all zeros for all classes, except class $k$ (which is unity). To make the GAN learn class-to-class translation, we propose to transform input images $x_A$ and $x_B$ to $x_A'=M^k_A\odot x_A$ and $x_B'=M^k_B\odot x_B$, where $\odot$ is the element-wise product. Next, we use these $x'$'s in the above losses. 

A problem with this scheme is that, while the network learns to transform individual classes using semantic dropout, it may miss learning the inter-class context within images. To this end, we propose to use the dropout stochastically, i.e., with a probability $p$, we select a label $k$ randomly from classes common to a pair of tuples $(x_A, g_A)$ and $(x_B, g_B)$. Next, using the respective ground truth masks, we create new semantic masks $M^A_k$ and $M^B_k$, which are then used to select the respective image pixels to generate $x'_A$ and $x'_B$ as described above. The full dropout pipeline is provided in Algorithm~\ref{alg:1}.

\begin{algorithm}  
	\SetAlgoLined
	\KwIn{tuples $(x_A, g_A)$ and $(x_B, g_B)$; dropout $p$}
    $l_A\setto get\_labels(g_A)$\tcp*[l]{get labels in $g_A$}
    $l_B \setto get\_labels(g_B)$\;
    $l_{AB} \setto l_A \cap l_B$\tcp*[l]{find common labels}
    \eIf{$\text{uniform}(0,1) <= p$}{
    	$k\follows \text{multinomial}(|l_{AB}|)$\tcp*[l]{$|\ .\ |$ is the set cardinality.}
        $M^k_A \setto get\_mask(l_{AB}[k], g_A)$\tcp*[l]{generate a mask on $g_A$ with $l_{AB}[k]$}
        $M^k_B \to get\_mask(l_{AB}[k], g_B)$\;
        $x'_A \to M^k_A\odot x_A$\; $x'_B \to M^k_B\odot x_B$\;
        \KwRet{$(x'_A, M_k^A\odot g_A)$, $(x'_B, M_k^B\odot g_B)$} \tcp*[l]{assuming $g$'s are 1-indexed.}
        }{
    	\KwRet{$(x_A, g_A)$, $(x_B, g_B)$}
    }
    \label{alg:1}
	\caption{Semantic dropout.}	
\end{algorithm}
 % It takes as input two unpaired tuples of image-ground-truths. With random probability $p$, the semantic classes are are dropped and the respective appearances in the images are removed. The updated images and ground truth labels are returned to be used in Sem-GAN as described in Figure~\ref{fig:arch}.

\vspace*{-0.7cm}
\section{Experiments}
\label{sec:expts}
We use three datasets and six image translation tasks to demonstrate the improvements afforded by Sem-GAN. Details of these datasets, tasks, network architectures, and our evaluation protocols follow. We also report results on improving the semantic segmentation accuracy.
\subsection{Datasets}
\vspace*{-0.7cm}
\noindent\paragraph{Cityscapes (CS)~\cite{cordts2016cityscapes}:} consists of 5K densely annotated real-world road scene images collected from 50 European cities and annotated for 30 semantic classes. The dataset has moderate diversity in weather and lighting conditions. % and contain several dynamic objects, including vehicles and pedestrians.  
%\vspace*{-0.5cm}
\noindent\paragraph{Mitsubishi Precision (MP):} consists of about 20K road scene images generated by the Mitsubishi Precision Co. simulator and densely annotated for 36 semantic classes. The dataset has high-resolution images from varied weather (summer, winter, rain), lighting conditions (dawn, dusk, night), and object appearances. %From a translation perspective, this dataset is challenging as the domain shift from the real images is significant. 
%\vspace*{-0.6cm}
\noindent\paragraph{VIPER~\cite{richter2017playing}:} (which is a recent version of the popular GTA5 dataset~\cite{richter2016playing}) consists of 250K frames from driving videos in realistic virtual worlds generated by the Unity gaming engine. The dataset is densely annotated for 31 semantic classes and includes images from varied weather and lighting conditions. %This dataset is more realistic than the MP dataset. % and thus the domain shift is lower.
\vspace*{-0.1cm}
\subsection{Data Preparation}
As the images in our datasets are of different resolutions, we resize them to a common size of 540$\times$ 860 pixels. Further, since the synthetic images are from video sequences, nearby frames might be very similar. To this end, we uniformly sample 5K frames from each of the synthetic datasets. We map the semantic classes from all the datasets to a common subset, with the Cityscapes annotations as the reference. We find that 19 classes are common, and use only these to enforce semantic consistency. Details are available in the supplementary material. We report experiments on five bi-directional translation tasks, namely (i) CS$\lftrt$MP, (ii) CS$\lftrt$Viper, (iii) CS Summer$\lftrt$ MP Winter, (iv) CS Day $\lftrt$ MP night, and (v) MP Summer$\lftrt$ MP Winter. We also present experiments on the task of mapping segmentation masks to real images (Seg $\to$ CS) to show that conditioning the generators directly on the segment labels may not be a replacement to our scheme. In this case, we use unpaired translations, that is we have sets of masks and images, without correspondences. Thus, our setting is different from that of pix2pix~\cite{isola2017image}. Note that, even though we use limited ground truth segmentation masks on the Cityscapes dataset, our problem remains unpaired as we do not assume correspondences between such image-label pairs across domains.

\subsection{Network Architectures}
We implemeted Sem-GAN from the code shared as part of Cycle-GAN\footnote{\tiny{\url{https://github.com/junyanz/pytorch-CycleGAN-and-pix2pix}}} using PyTorch~\cite{paszke2017automatic}. For the generators and discriminators in GAN, we use a sequence of 9 residual network blocks. The Adam optimizer~\cite{kingma2014adam} is used for training the networks, with an initial learning rate of $0.0002$. The validation accuracy seem to saturate in about 50 epochs for all our tasks, except for the Seg$\to$CS task, for which we use 200 epochs. For our segmentation networks, we use the FCN~\cite{long2015fully} implementation in Pytorch.\footnote{\tiny{\url{https://github.com/ZijunDeng/pytorch-semantic-segmentation}}}, which is cheaper and faster to train alongside other modules in our framework in comparison to other deeper networks such as Deeplab~\cite{chen2016deeplab} and PSP-Net~\cite{zhao2017pyramid}. In FCN, we use a VGG-16 network and use cross-entropy loss on the final output layer for enforcing the Sem-GAN criteria.

\subsection{Training, Testing, and Evaluation}
We define training, validation, and test sets by random sampling each dataset in 85:5:10 splits. The images are cropped to $256\times 256$ pixels; the training inputs are cropped randomly during training (as part of data augmentation), while the validation and test images are center-cropped. The segmentation networks are pre-trained on the respective training sets to recognize 19 semantic classes. Note that, we use only images from ideal weather conditions (well-lit and good weather) for this training, while networks for other conditions (day, night, winter, etc.) are learned jointly with other modules in Sem-GAN. For training the segmenters, we fine-tune a VGG-16 model with batches of 16 images, optimizing the parameters using stochastic gradient descent with a learning rate of $0.0001$ and a momentum of 0.9. During testing, we do not use the segmentation pipeline, instead directly forward pass the source images through the generators and gather the translated images for evaluation.

For quantitative evaluations, we use the semantic segmentation accuracy of the translated images by a segmentation model trained on the respective domain. To ensure unbiased evaluation, we report results using two segmentation networks, namely (i) FCN and (ii) PSP-Net~\cite{zhao2017pyramid}. The evaluation networks are trained separately from Sem-GAN on training sets from the respective domains. Using these models, we report results on (i) the overall accuracy (Over. Acc) -- which is the number of pixels correctly predicted against the total number of annotated pixels, (ii) the average class accuracy (Avg. Acc) -- which is the average of per-class accuracy, and (iii) the mean intersection-over-union (mIoU) score~\cite{long2015fully} over all the classes. On the 19 evaluation classes, FCN achieves mIoU of 64.1\%, 56.2\%,	and 51.7\% on the test sets of MP, Viper, and CS datasets respectively, while PSP-net gets 73.4\%, 71.1\% and 61.1\%. We use 1K images randomly sampled from the Cityscapes dataset for training the respective segmentation models on this dataset.

\subsection{Semantic Dropout}
\begin{figure*}[h]
\centering
%\subfigure[MP$\leftrightarrow$CS]{\label{fig:inc_p}\includegraphics[width=4cm,trim={0.2cm 7cm 1cm 7.5cm},clip]{figs/inc_p.pdf}}
\subfigure[CS$\to$MP]{\label{fig:inc_iter_AB}\includegraphics[width=4.3cm,trim={0.2cm 7cm 1cm 7.5cm},clip]{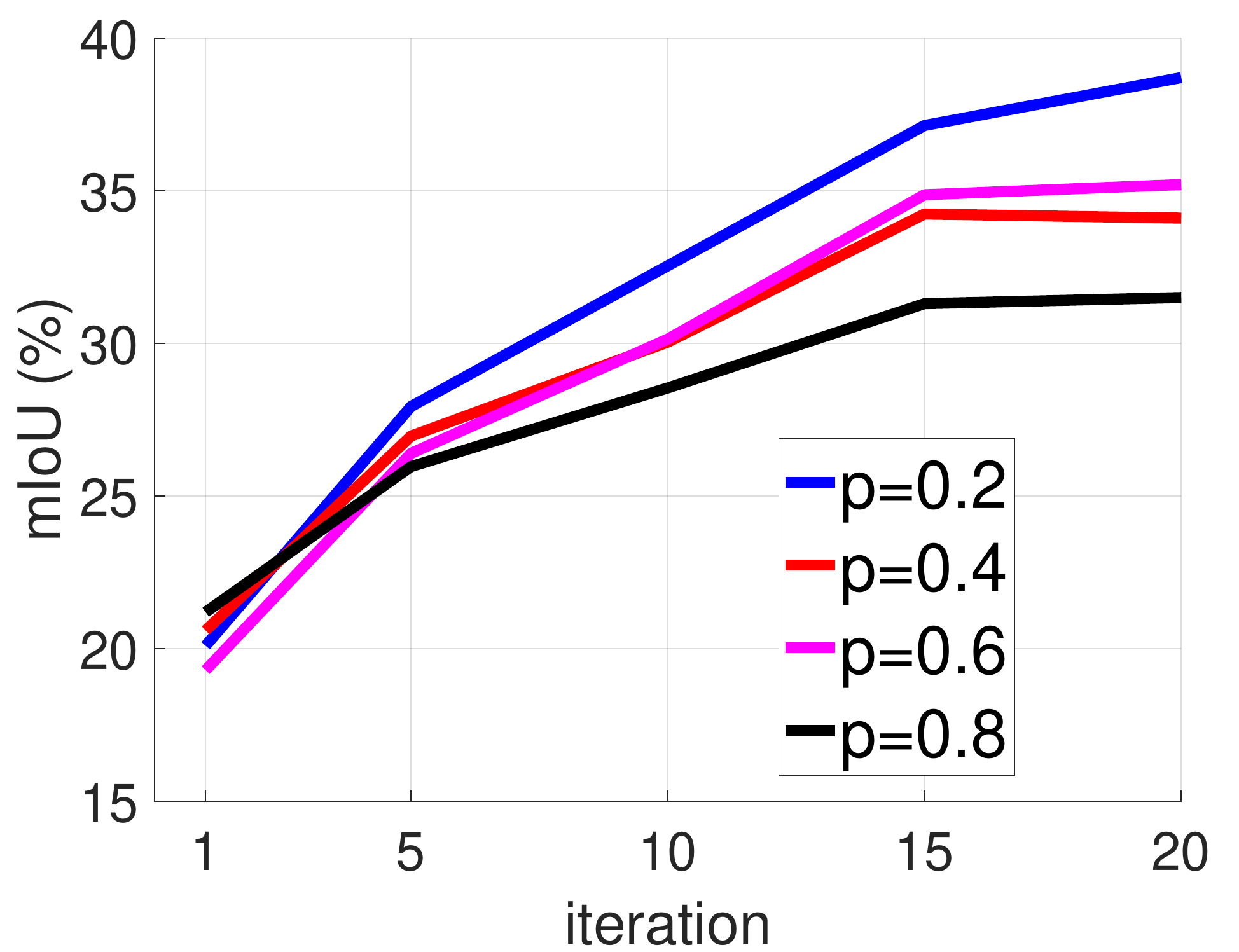}}
\subfigure[MP$\to$CS]{\label{fig:inc_iter_BA}\includegraphics[width=4.3cm,trim={0.2cm 7cm 1cm 7.5cm},clip]{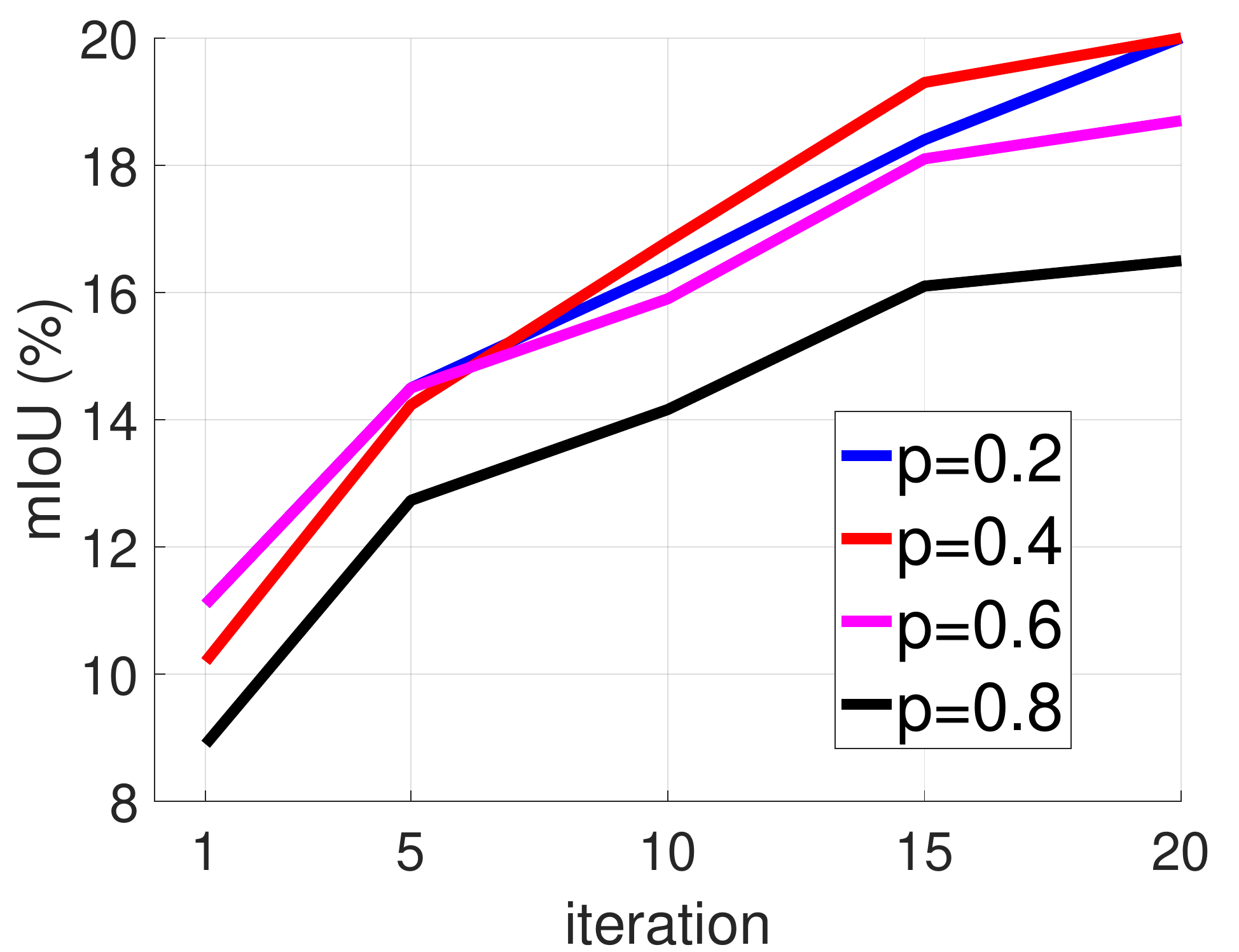}}
\subfigure[MP$\to$CS]{\label{fig:mms2cs_A2B}\includegraphics[width=4.3cm,trim={0.5cm 7cm 0.5cm 7.5cm},clip]{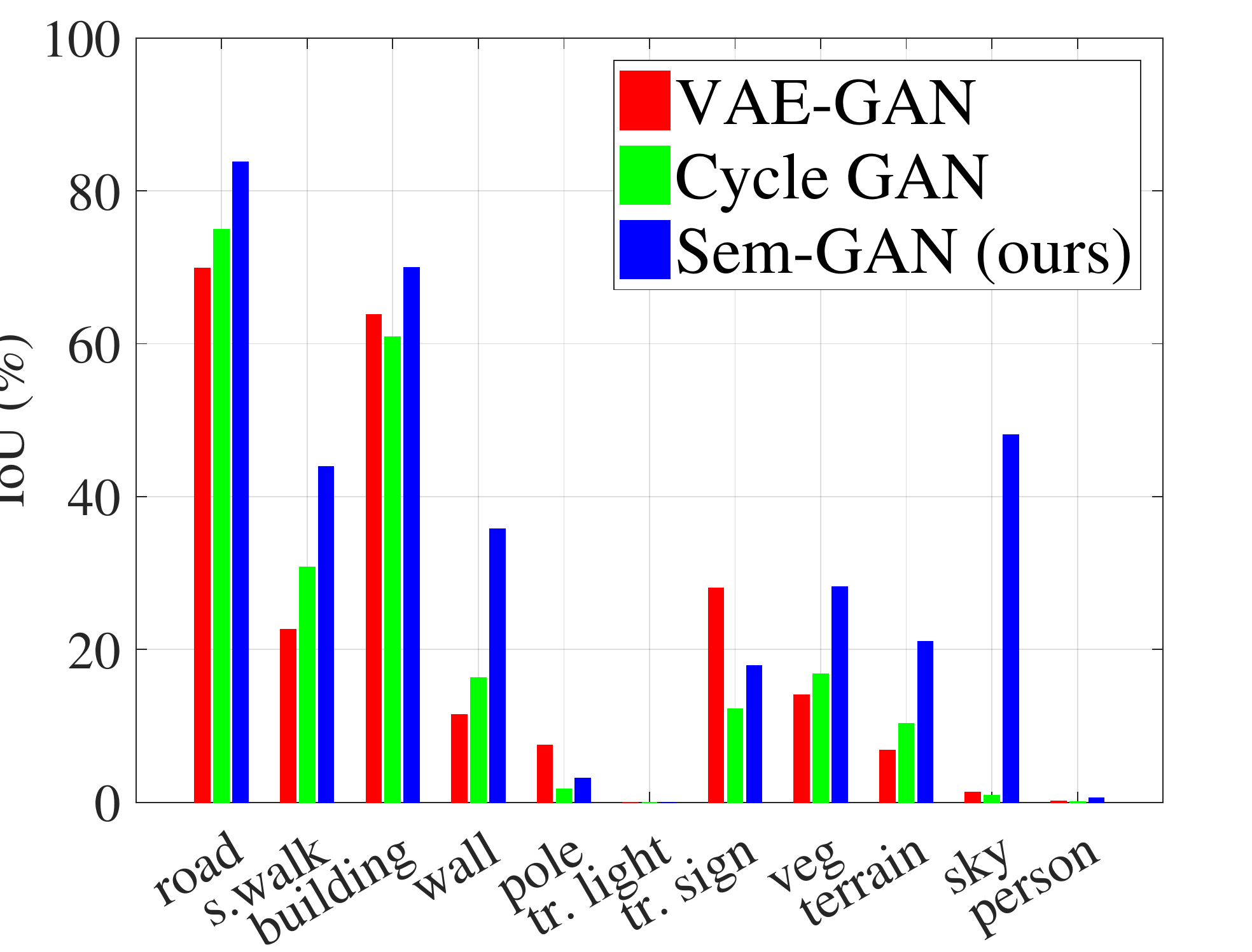}}
\subfigure[VP$\to$CS]{\label{fig:viper2cs_A2B}\includegraphics[width=4.3cm,trim={0.5cm 7cm 0.5cm 7.5cm},clip]{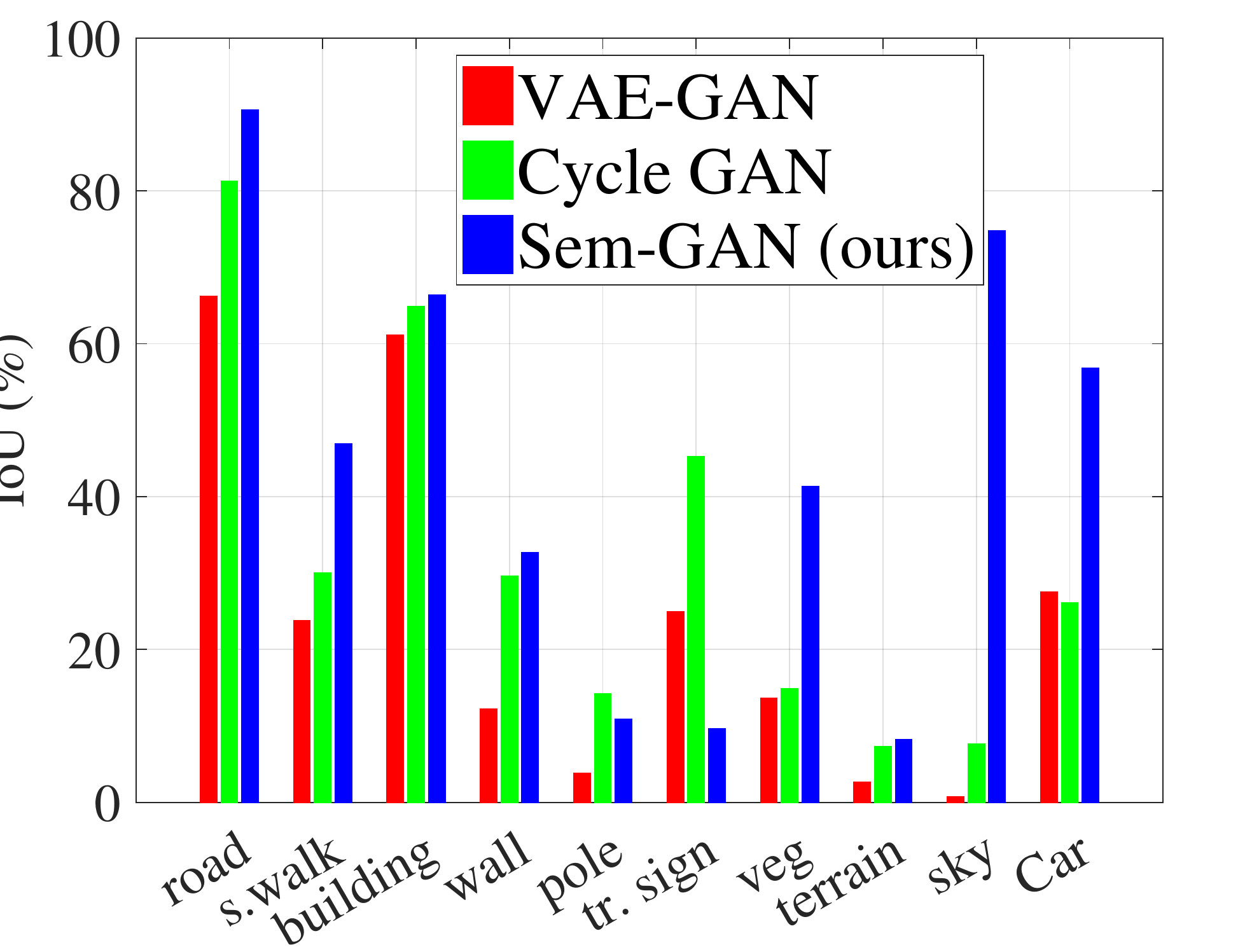}}
\caption{Effect of semantic dropout on image translation accuracy. Right: Detailed analysis of accuracies when translating each class.}
\label{fig:eval_smdrop}
\end{figure*}

%An important idea enabled by our framework is the use of semantic dropout, described in Section~\ref{sec:smdrop}, which uses a stochastic scheme to remove segments randomly from the images. Our hypothesis is that such a scheme might allow learning segment-to-segment translations specifically improving the convergence and translation performance. In this section, we evaluate this idea quantitatively. 
We first analyze the merit of semantic-dropout. This scheme has a parameter $p$, which is the probability of dropping segments; a higher-value of $p$ drops segments too frequently; as a result, the generators may not be able to learn their spatial contexts. To this end, in Figures~\ref{fig:inc_iter_AB} and~\ref{fig:inc_iter_BA}, we plot the mIoU for the MP$\leftrightarrow$CS tasks against varying $p=\set{0, 0.2, 0.4, 0.6, 0.8}$. As is clear from the plots, semantic dropout improves the performance of translation significantly; e.g., on the CS$\to$MP task, the gap between $p=0$ and $p=0.2$ is $\sim$20\%. We also see that a higher value $p=0.8$ shows lower accuracy. Thus, we use $p=0.2$ for CS$\to$MP and $p=0.4$ for MP$\to$CS and Viper$\to$CS.

\begin{table}
	\centering
	\begin{adjustbox}{width=1\linewidth}
		\renewcommand{\arraystretch}{1}
		\begin{tabular}{c|c|c|c|c||c|c|c}
			A$\lftrt$B &\multicolumn{1}{c}{}&\multicolumn{3}{|c||}{A$\to$B} & \multicolumn{3}{|c}{B$\to$A}\\
			\hline
			Task & Scheme & Avg. Acc & Over. Acc& mIoU & Avg. Acc& Over. Acc & mIoU\\
			\hline
			%MP$\to$CS& \multicolumn{6}{c}{}\\
			\hline
			MP$\lftrt$CS&VAE-GAN    & 42.6	 & 59.4	  & 13.2    & 13.2	 & 30.9	& 6.1 \\
			&Style-Trans. & 44.5 & 	72.1	& 26.3 &  15.6 & 27.8 &	 6.3\\
			&Cycle-GAN        & 36.7    & 56.2   & 16.2 & 19.5 & 36.9 & 7.2\\
			&Sem-GAN  & 51.5    & 71.9   & 34.1 & 29.1 & 58.4 & 18.3\\
			&Sem-GAN+SM  & 60.7	 & 80.2	  & \textbf{40.2} & 29.4 &	67.8 & \textbf{19.3}\\
			\hline
			%Viper$\to$CS     &  \multicolumn{6}{c}{}\\
			%\hline
			Viper$\lftrt$CS&VAE-GAN &    31.3	  & 54.5   &   13.4 &  18.9	 & 36.4	 & 8.6 \\
			& Style Trans. & 29.3 & 	64.7 & 	13.7 & 17.9	& 61.6 & 11.0\\
			&Cycle-GAN        & 23.6    & 54.1   & {9.2} &  21.0  &  63.9 &  13.4\\
			&Sem-GAN   & 38.8    & 82.0   &   24.0 &  27.4  &  80.3 &  {20.4}\\
			&Sem-GAN+SM& 42.5	  & 84.2   & \textbf{28.4} & 27.7 &	81.6 &	\textbf{21.5}\\
			\hline
			%MP(N)$\to$CS(D)       &\multicolumn{6}{c}{} \\
			%\hline
			MP(N)$\lftrt$CS(D) &VAE-GAN  &  22.6	  & 39.6   &  7.4   &  10.7	 & 20.0	 &  5.5\\
			&Cycle-GAN        & 32.8    & 47.5   & 10.7 &  14.8  &  48.2 &  7.1\\
			&Sem-GAN   & 54.1    & 79.7   & {32.7} &  27.6  &  78.3 &  \textbf{20.5}\\
			&Sem-GAN+SM& 56.2	  & 80.0   & \textbf{36.7} & 28.6	& 78. 1	 & 20.2 \\
			\hline
			%MP(W)$\to$CS(S)       & \multicolumn{6}{c}{}\\
			%\hline
			MP(W)$\lftrt$CS(S) &VAE-GAN    & 26.2	  & 48.5   &  8.0   &  9.08	 & 41.1  &	4.7\\
			&Cycle-GAN        & 27.1    & 65.9   & 13.2 &  12.8  &  51.9 &  7.1\\
			&Sem-GAN    & 51.3    & 85.9   & {32.3} &  22.4  &  76.4 &  {16.2}\\
			&Sem-GAN+SM& 50.1	  & 84.2   & \textbf{34.2} & 22.5	& 72.1	 & \textbf{16.9}\\
			\hline
			%MP(S)$\to$MP(W)       &\multicolumn{6}{c}{} \\
			%\hline
			MP(S)$\lftrt$MP(W) &VAE-GAN  &  53.0	 & 87.1	  & 41.8 &  57.6  &  75.9 & 45.2\\
			&Cycle-GAN        & 60.0    & 74.6   & {47.4} &  61.8  &  91.0 &  51.0\\
			&Sem-GAN   & 53.1    & 75.9   & 43.4 &  62.9  &  92.2 & {52.3}\\
			&Sem-GAN+SM& 54.7	 & 75.1	  & \textbf{45.5} &  63.2 &	92.3 &	\textbf{53.3}\\
			\hline
			%Seg$\to$CS       & \multicolumn{6}{c}{} \\
			\hline
			Seg$\to$CS &VAE-GAN    & 7.36	  & 48.0   & 4.0   & NA & NA & NA\\ 
			&Cycle-GAN        & 12.6    & 37.7   & 7.8 &  NA  &  NA &  NA\\
			&Sem-GAN    & 35.6    & 75.0   & \textbf{26.6} &  NA  &  NA &  NA\\
		\end{tabular}
	\end{adjustbox}
	\caption{Results using our Sem-GAN and semantic dropout (SM) against the state-of-the-art Cycle-GAN~\cite{zhu2017unpaired}, VAE-GAN~\cite{liu2017unsupervised}, and style transfer model~\cite{johnson2016perceptual}. We use the FCN~\cite{long2015fully} for evaluation. All numbers are in \%. W=Winter, S=Summer, D=Day, and N=Night.}
	\label{tab:soa}
\end{table}

\begin{table}
	\centering
	\begin{adjustbox}{width=1\linewidth}
		\renewcommand{\arraystretch}{1}
		\begin{tabular}{c|c|c|c|c||c|c|c}
			A$\lftrt$B &\multicolumn{1}{c}{}&\multicolumn{3}{|c||}{A$\to$B} & \multicolumn{3}{|c}{B$\to$A}\\
			\hline
			Task & Scheme & Avg. Acc & Over. Acc& mIoU & Avg. Acc& Over. Acc & mIoU\\
			\hline
			%MP$\to$CS& \multicolumn{6}{c}{}\\
			\hline
			MP$\lftrt$CS&Cycle-GAN & 45.6 & 	62.6 & 	21.5	&	23.7 &	55.9 &	11.6\\
			&Sem-GAN & 50.0	& 77.4	 & \textbf{34.2}	&	30.1	 & 71.2	 & \textbf{18.6}\\
			\hline
			%Viper$\to$CS     &  \multicolumn{6}{c}{}\\
			%\hline
			Viper$\lftrt$CS&Cycle-GAN & 29.8	 & 57.3	 & 15.4	   & 	24.6 &  70.7 & 17.3\\
			&Sem-GAN & 38.3	 &74.0   & 	\textbf{23.2}   &  30.9	 & 80.8 & 	\textbf{23.5}\\
			\hline
			%MP(N)$\to$CS(D)       &\multicolumn{6}{c}{} \\
			%\hline
			MP(N)$\lftrt$CS(D) &Cycle-GAN & 34.1	 & 48.2  & 	15.5 & 19.0	 & 57.2  &	8.97\\
			&Sem-GAN & 49.8 & 	74.0  & 	\textbf{27.9} & 	29.3 & 	77.5	 & \textbf{20.2}\\
			\hline
			%MP(W)$\to$CS(S)       & \multicolumn{6}{c}{}\\
			%\hline
			MP(W)$\lftrt$CS(S) &Cycle-GAN & 33.9 & 	63.3	 & 14.8 & 		13.4 & 	52.1	 & 7.39\\
			&Sem-GAN  & 48.3	 & 76.8	   & \textbf{26.2}   & 		23.5	 & 75.8	 & \textbf{16.9}\\
			\hline
			%MP(S)$\to$MP(W)       &\multicolumn{6}{c}{} \\
			%\hline
			MP(S)$\lftrt$MP(W) &Sem-GAN & 64.6 & 	76.7	& 51.1 & 		65.6	 & 91.6 & 	\textbf{54.8} \\
			&Sem-GAN & 57.9	 & 76.7	  & \textbf{57.2} &  63.2  & 90.7 & 	{51.9} \\
			\hline
			%Seg$\to$CS       & \multicolumn{6}{c}{} \\
			\hline
			Seg$\to$CS&Cycle-GAN & 16.2	 & 46.0  & 9.92 & NA & NA & NA\\
			&Sem-GAN & 19.7	 & 55.0 & 	\textbf{13.4} &  NA  &  NA &  NA\\
		\end{tabular}
	\end{adjustbox}
	\caption{Comparisons between Sem-GAN and Cycle-GAN~\cite{zhu2017unpaired} using PSP-net~\cite{zhao2017pyramid} segmentation model for evaluation. W=Winter, S=Summer, D=Day, and N=Night.}
	\label{tab:soa-p}
\end{table}

\begin{table*}[h]
	\centering
	\begin{adjustbox}{width=1\linewidth}
		\renewcommand{\arraystretch}{1} 
		\begin{tabular}{|l|l|l|l|l|l|l|l|l|l|l|l|l|l|l|l|l|l|l|l|l|c|c|c}
			\hline
			Method &\rb{n/w} & \rb{Road} & \rb{s.walk} & \rb{Bldg} & \rb{wall} & \rb{fence} & \rb{pole} & \rb{t. light} & \rb{t. sign} & \rb{veg.} & \rb{terrain} & \rb{sky} & \rb{person} & \rb{rider} &  \rb{car} & \rb{truck} & \rb{bus} & \rb{train} & \rb{m.cycle} & \rb{bicycle}& \tb{mIoU} &  {Ov. Acc} \\
			\hline
			CS only & \tb{FCN} & 85.1 & 38.9 & 60.6 & 0.8 & 1.1 & 0.0 & 0.0 & 0.0 & 65.1 & 7.4 & 36.3 & 19.8 & 0.0 & 62.7 & 0.0 & 0.0 & 0.0 & 0.0 & 0.4 &  19.9 &   77.4 \\
			\hline
			% MP only & F\\
			% SemGAN(MP) & F \\
			CS + MP & FCN & 87.0 & 41.8 & 64.6 & 14.5 & 0.2 & 0.6 & 0.1 & 0.4 & 68.6 & 11.7 & 67.3 & 11.2 & 0.0 & 63.6 & 0.0 & 0.0 & 0.0 & 0.0 & 0.1 &  22.7 &    79.3 \\
			CS+Cy(MP) & FCN & 85.5 & 40.3 & 63.6 & 6.9 & 0.0 & 3.2 & 0.4 & 3.5 & 69.0 & 7.8 & 52.2 & 11.7 & 0.0 & 62.5 & 0.0 & 0.0 & 0.0 & 0.0 & 2.6 &  21.5 &   77.8  \\
			CS+Sm(MP) & FCN & 88.1 & 47.0 & 67.8 & 12.8 & 0.5 & 7.1 & 0.0 & 2.0 & 71.1 & 10.0 & 69.0 & 15.4 & 0.0 & 67.6 & 0.0 & 0.0 & 0.0 & 0.0 & 4.0 &  \tb{24.3} &   80.8 \\
			\hline
			% VP only & F 
			% SemGAN(VP) & F \\
			CS + VP & FCN & 90.2 & 54.1 & 70.3 & 22.6 & 5.9 & 3.4 & 0.0 & 3.4 & 73.4 & 27.2 & 67.1 & 31.4 & 0.3 & 73.2 & 9.1 & 18.7 & 5.6 & 0.2 & 21.2 &  30.4 &    83.2  \\ 
			CS+Cy(VP) & FCN & 90.3 & 53.2 & 67.4 & 9.4 & 15.3 & 2.6 & 0.1 & 4.8 & 70.6 & 26.4 & 60.9 & 36.4 & 0.9 & 74.2 & 21.1 & 24.9 & 3.4 & 5.0 & 30.4 &  31.4  &  82.8 \\
			CS+Sm(VP) &FCN & 92.1 & 59.1 & 71.3 & 21.6 & 19.1 & 4.4 & 0.2 & 5.6 & 74.1 & 30.2 & 70.1 & 36.4 & 1.3 & 76.8 & 24.2 & 20.5 & 11.7 & 4.3 & 30.7 &  \tb{34.4} &  84.6 \\
			\hline
			CS only & \tb{PSP} & 85.2 & 35.2 & 62.9 & 4.1 & \tb{15.4} & 0.5 & 0.0 & 2.6 & 68.6 & 24.3 & 49.0 & 27.2 & 0.0 & 63.6 & 0.0 & 4.8 & 0.0 & 0.0 & 19.7 &  24.4 &  78.7 \\
			\hline
			%MP only & PSP & 51.2 & 12.9 & 40.7 & 4.6 & 0.0 & 4.8 & 0.2 & 9.3 & 50.5 & 2.6 & 10.3 & 0.0 & 0.0 & 59.5 & 0.0 & 0.0 & 0.0 & 0.0 & 0.0 &  13.0 &  56.0 &  42.0 \\
			%Cy(MP) & P & 83.2 & 32.2 & 49.3 & 7.3 & 0.0 & 5.1 & 0.8 & 14.0 & 43.9 & 10.0 & 28.1 & 0.0 & 0.0 & 50.5 & 0.0 & 0.0 & 0.0 & 0.0 & 0.0 &  17.1 &    69.1 &  56.4 \\
			%Sm(MP) & P & 85.4 & 40.9 & 63.8 & 11.9 & 0.0 & 8.5 & 1.2 & 10.0 & 69.9 & 13.9 & 64.0 & 0.0 & 0.0 & 63.2 & 0.0 & 0.0 & 0.0 & 0.0 & 0.0 &  \tb{22.8} &    78.2 &  66.0\\
			\hline
			CS+MP &PSP & 88.8 & 49.5 & 70.2 & 7.0 & 4.7 & 9.0 & 0.0 & 13.4 & 72.8 & 20.2 & 74.9 & 38.2 & 0.0 & 73.5 & 0.0 & 3.2 & 0.0 & 0.0 & 31.5 &  29.3 &    82.4 \\
			%CS+Cy(MP) & PSP & 89.8 & 51.3 & 71.3 & 14.1 & 4.2 & 11.2 & 0.7 & 17.9 & 73.3 & 23.7 & 63.5 & 39.2 & 0.7 & 73.2 & 0.0 & 7.6 & 0.0 & 0.0 & 34.7 &  30.3 &   72.6 \\
			CS+Sm(MP) &PSP & 90.3 & 54.2 & 72.4 & 17.4 & 8.0 & 16.6 & 0.1 & 17.9 & 75.8 & 23.6 & 74.2 & 42.8 & 8.5 & 74.3 & 0.0 & 17.3 & 0.0 & 0.0 & 36.1 &  \tb{33.1} &    84.3 \\
			\hline
			%VP only & PSP & 75.9 & 29.4 & 49.4 & 0.0 & 0.0 & 0.0 & 0.6 & 10.8 & 65.2 & 13.2 & 62.5 & 15.7 & 0.0 & 58.3 & 12.7 & 6.2 & 0.0 & 0.1 & 0.0 & 21.1 & 71.8 & 57.8\\
			%Cy(VP) & P & 85.2 & 35.2 & 53.6 & 0.0 & 4.7 & 0.0 & 4.3 & 9.6 & 22.3 & 15.3 & 20.3 & 11.3 & 0.0 & 66.4 & 11.8 & 3.0 & 0.0 & 4.1 & 0.0 &  18.3 &    70.9 &  56.8 \\
			%Sm(VP) & P & 87.9 & 37.0 & 56.5 & 0.0 & 4.5 & 0.0 & 2.2 & 15.6 & 42.6 & 24.8 & 40.4 & 20.0 & 0.0 & 74.1 & 19.9 & 14.7 & 0.0 & 4.6 & 0.0 &  \tb{23.4} &    76.2 &  62.8 \\
			\hline
			CS+VP & PSP & 91.5 & 54.2 & 74.8 & 23.8 & 7.4 & 18.3 & 3.0 & 13.7 & 76.9 & 24.8 & 66.5 & 48.6 & 22.2 & 82.1 & 35.7 & 19.7 & 28.2 & 6.4 & 42.7 &  39.0 &    85.6 \\
			CS+Sm(VP) & PSP & \tb{93.4} & \tb{63.4} & \tb{76.4} & \tb{27.3} & 11.7 & \tb{23.6} & \tb{15.8} & \tb{23.6} & \tb{78.2} & \tb{32.0} & \tb{78.4} & \tb{52.2} & \tb{26.2} & \tb{84.0} & \tb{33.4} & \tb{33.4} & \tb{30.1} & \tb{18.1} & \tb{42.4} &  \tb{44.4} &    \tb{87.4} \\ 
			% CS+CyGAN(VP) & P\\
			\hline
		\end{tabular}
	\end{adjustbox}
	\caption{Training segmentation models using adapted images. We adapt synthetic MP and VIPER (VP) datasets to Cityscapes (CS) domain. Sm(MP) and Cy(MP) denote adaptation of all images in MP with our Sem-GAN and Cycle GAN respectively to the CS domain. "only" refers to using images directly from that domain (without adaptation). We use two segmentation CNNs: "FCN" is VGG-FCN8s and "PSP" is PSPNet for segmentation evaluation. We also show per-class accuracy on the 19 common classes across the three datasets.}
	\label{tab:seg_acc}
\end{table*}

%\begin{figure}
%\centering
%\subfigure[MP$\to$CS]{\label{fig:mms2cs_A2B}\includegraphics[width=5.5cm,trim={0.5cm 7cm 0.5cm 7.5cm},clip]{figs/compare_classes_mms2cs_A2B.pdf}}
%\subfigure[CS$\to$MP]{\label{fig:mms2cs_B2A}\includegraphics[width=5.5cm,trim={0.5cm 7cm 0.5cm 7.5cm},clip]{figs/compare_classes_mms2cs_B2A.pdf}}
%\caption{Detailed analysis of accuracies when translating each class. We show only 9 classes for which there are sufficient number of images in the test set.}
%\label{fig:barplot}
%\vspace*{-0.5cm}
%\end{figure}
%

\subsection{State-of-the-Art Comparisons}
\begin{figure*}[!h]
	\centering
	\subfigure[CS (orig)]{\includegraphics[width=2.8cm]{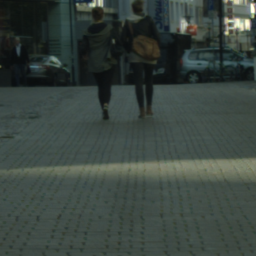}}
	\subfigure[CS$\to$ MP Cy-GAN]{\includegraphics[width=2.8cm]{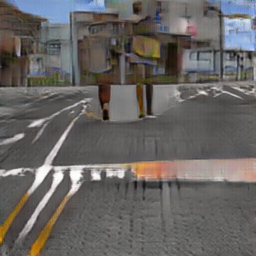}}
	\subfigure[CS$\to$ MP Sem-GAN]{\includegraphics[width=2.8cm]{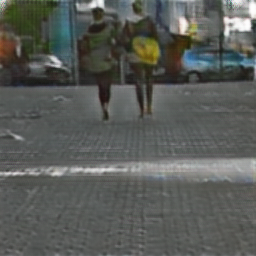}}
	\vspace*{-0.2cm}
	\subfigure[VP (orig)]{\includegraphics[width=2.8cm]{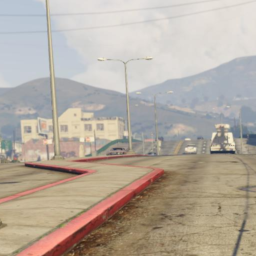}}
	\subfigure[VP$\to$CS  Cy-GAN]{\includegraphics[width=2.8cm]{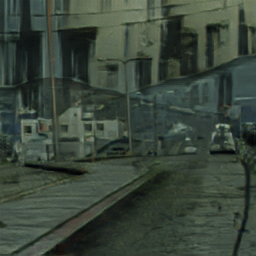}}
	\subfigure[VP$\to$CS Sem-GAN]{\includegraphics[width=2.8cm]{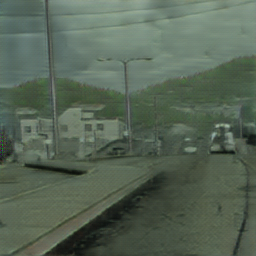}}\\
	%\caption{Synthetic (Viper) to Cityscapes translation.}
	\vspace*{-0.1cm}
	\subfigure[Segment (M)ask]{\includegraphics[width=2.8cm]{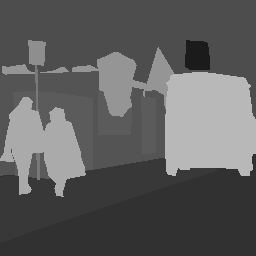}}
	\subfigure[M$\to$CS Cy-GAN]{\includegraphics[width=2.8cm]{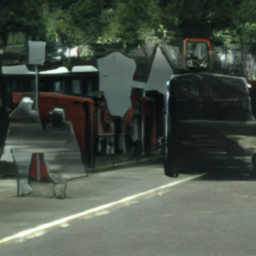}}
	\subfigure[M$\to$CS Sem-GAN]{\includegraphics[width=2.8cm]{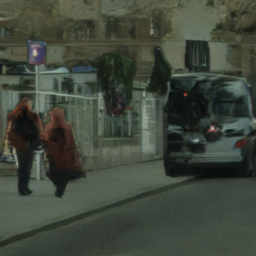}}
	\vspace*{-0.1cm}
	\hspace*{0.5cm}\subfigure[MP (orig)]{\includegraphics[width=2.8cm]{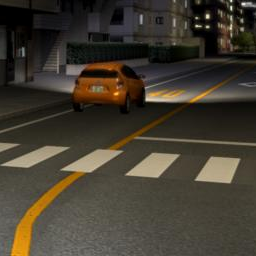}}
	\subfigure[MP$\to$CS]{\includegraphics[width=2.8cm]{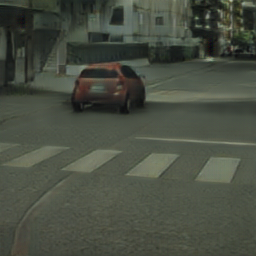}}\\
	\subfigure[CS (orig)]{\includegraphics[width=2.8cm]{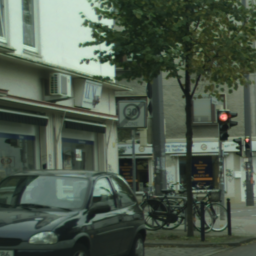}}
	\subfigure[CS$\to$MP]{\includegraphics[width=2.8cm]{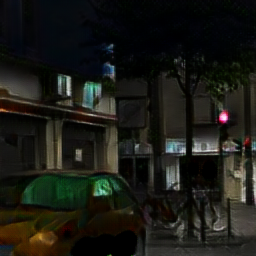}}
	\subfigure[MP (orig)]{\includegraphics[width=2.8cm]{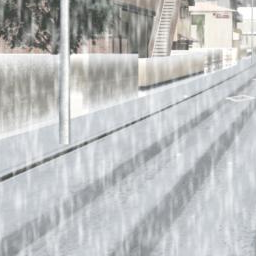}}
	\subfigure[MP$\to$CS ]{\includegraphics[width=2.8cm]{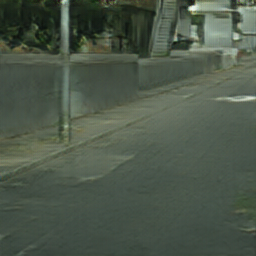}}
	\subfigure[CS (orig)]{\includegraphics[width=2.8cm]{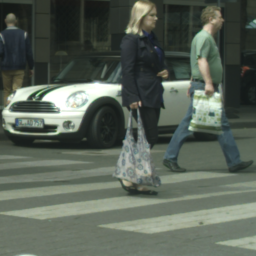}}
	\subfigure[CS$\to$MP]{\includegraphics[width=2.8cm]{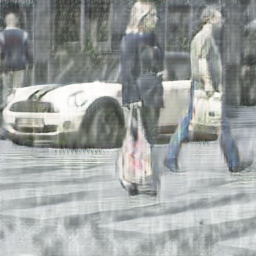}}
	\caption{Qualitative results; (a--c) CS$\to$ MP, (d--f)  Viper$\to$CS, (g--i) segment mask$\to$ CS, (j--m)  Night$\leftrightarrow$Day, and (n--q) Winter$\leftrightarrow$Summer. More results in the supplementary material.}
	\label{fig:qualitative_results}
\end{figure*}

In Tables~\ref{tab:soa} and~\ref{tab:soa-p}, we compare Sem-GAN against three state-of-the-art image translators: (i) Cycle-GAN~\cite{zhu2017unpaired}, (ii) VAE-GAN~\cite{liu2017unsupervised}, and style-transfer using perceptual losses~\cite{johnson2016perceptual}. We also report performances with and without semantic dropout (SM). As is clear, Sem-GAN (+ SM) outperforms Cycle-GAN in almost all tasks, especially on the challenging mIoU criteria. Specifically, we find that on MP$\to$CS and Viper$\to$CS tasks, our scheme is nearly 20\% better in classification accuracies. Similar results are observed on other tasks as well, except in MP(Snow)$\to$MP(Winter) translations. In this case, the source and target domains are inherently the same, except for simulated snow in the latter, which can be undone by the generator, thereby perfectly aligning the domains. In Figures~\ref{fig:viper2cs_A2B} and~\ref{fig:mms2cs_A2B}, we analyze the per-class IoU for the MP$\leftrightarrow$CS task. Note that not all classes are present in our (randomly chosen) test set. We see that Sem-GAN almost always shows superior translations on most classes. In Figure~\ref{fig:qualitative_results}, qualitative results are presented. On the Mask$\to$CS task, Sem-GAN guides the error from the segmenters to be improve the appearance of the generated segments as demonstrated by results in Tables~\ref{tab:soa} and~\ref{tab:soa-p}, leading to better results than the other models.

%Interestingly, we find that on CS$\to$MP translation, Sem-GAN performs the worst on the `rider' class, while other two schemes perform reasonably well; on investigation we found that Sem-GAN confuses `rider' with `person'.  %More results are in the supplementary material.
% that have sufficient number of test images to provide a statistically useful comparison
% In this case, we see that the performance of all the schemes are nearly the same. This is unsurprising as the domain shift in this case is only due to a holistic (non-linear) image transformation; which when undone by the generator, the domains can be perfectly aligned.
%An observation from the Table~\ref{tab:soa} is the apparent skew in the accuracies against the translation directions. For example, the results in the B$\to$A task are much lower compared to those on A$\to$B. When A and B are the synthetic and real domains respectively, this skew is perhaps because of the need for the generator to learn few-to-many mappings (as in A$\to$B) against many-to-few mappings (in B$\to$A), the latter being harder as object appearances may vary (even subtly) from frame to frame in real world images. However, for synthetic to real, the generator could learn to produce noise to make the synthetic objects appear as real to the discriminator, which is much easier to learn. 
 
%A task that needs to be highlighted is (unpaired) seg$\to$CS in Table~\ref{tab:soa}. As noted above, it may seem that conditioning on the semantic labels (source) may be sufficient to enforce semantic consistency (instead of our feedback scheme). However, note that even if the generators are conditioned on the class labels, the generated images for the respective classes may not be structurally valid, and thus may be unrecognizable to a segmenter (see e.g., Figure~\ref{fig:qualitative_results}(g--i)). Instead Sem-GAN guides the error from the segmenters to be improve the appearance of the generated segments as demonstrated by results in Tables~\ref{tab:soa} and~\ref{tab:soa-p}.

\subsection{Improvements on Semantic Segmentation}
Next, we analyze the merit of Sem-GAN for improving the original task, namely training semantic segmentation models via synthetic data. Our analysis is loosely based on~\cite{hoffman2017cycada}, however using our datasets and evaluation models. We use 10K images from the two synthetic datasets and 200 images from the Cityscapes (CS) dataset. We translated the synthetic images (source) to the CS domain (as in~\cite{hoffman2017cycada}) and used source ground truth labels for training two segmentation models. We used a test set of 500 CS images for evaluating our models. All the algorithms are trained using SGD with a learning rate of 0.0001 for 50 epochs. As is clear from Table~\ref{tab:seg_acc}, Cycle GAN is sometimes seen to reduce the performance (e.g., Cy(VP)) against no adaptation likely due to the correspondence mismatch problems alluded to earlier. However, Sem-GAN improves image adaptation significantly compared to CycleGAN, and leads to more accurate segmentation models than when not using adaptation; e.g., "CS only" with FCN8s is 19.9\% mIoU, while using Sem-GAN i.e., CS+Sem(VP), this improves to 34.4\%. Similarly, using PSPNet, "CS only" to CS+Sem(VP) is 24.4\% to 44.4\%, a 20\% improvement. Further, note that the improvement from CS+VP to CS+Sem(VP) is nearly 6\%; the former is without any adaptation on VP images. More comparisons and results are available in the Supplementary material. The code for the paper will be made publicly available.

\section{Conclusions}
\label{sec:conclude}
We presented an image-to-image translation framework that uses semantic consistency using segment class identities for achieving realistic translations. Modeling such a consistency as a novel loss, we presented an end-to-end learnable GAN architecture. We demonstrated the advantages of our framework on three datasets and six translation tasks. Our results clearly demonstrate that semantic consistency, as is proposed in this paper, is very important for ensuring the quality of the translation. %With better and faster segmentation algorithms, the quality of our translation scheme could be further improved.
%We presented our objective formulations with in standard GAN setup and provided an end-to-end learnable architecture. Our experiments on a variety of translation tasks demonstrated that using semantic consistency significantly improved the performance of these tasks. 

{\small
\bibliographystyle{ieee}
\bibliography{segcyclegan}
}
\appendix
%In this supplementary material, we provide:
%\begin{enumerate}
%	\item Additional results on the tasks.
%	\item Ablative analysis.
%	\item Qualitative results on tasks that do not have ground truth segmentation masks.
%	\item Qualitative results from various translation tasks.
%\end{enumerate}
\section*{Appendix}
\section{Additional Details and Comparisons}
\begin{figure}[h]
	\centering
	\subfigure[Viper$\to$CS]{\includegraphics[width=7.5cm,trim={0.5cm 7cm 0.5cm 7.5cm},clip]{figs/viper2cs_compare_classes_mms2cs_A2B_suppl.pdf}}
	\subfigure[CS$\to$Viper]{\includegraphics[width=7.5cm,trim={0.5cm 7cm 0.5cm 7.5cm},clip]{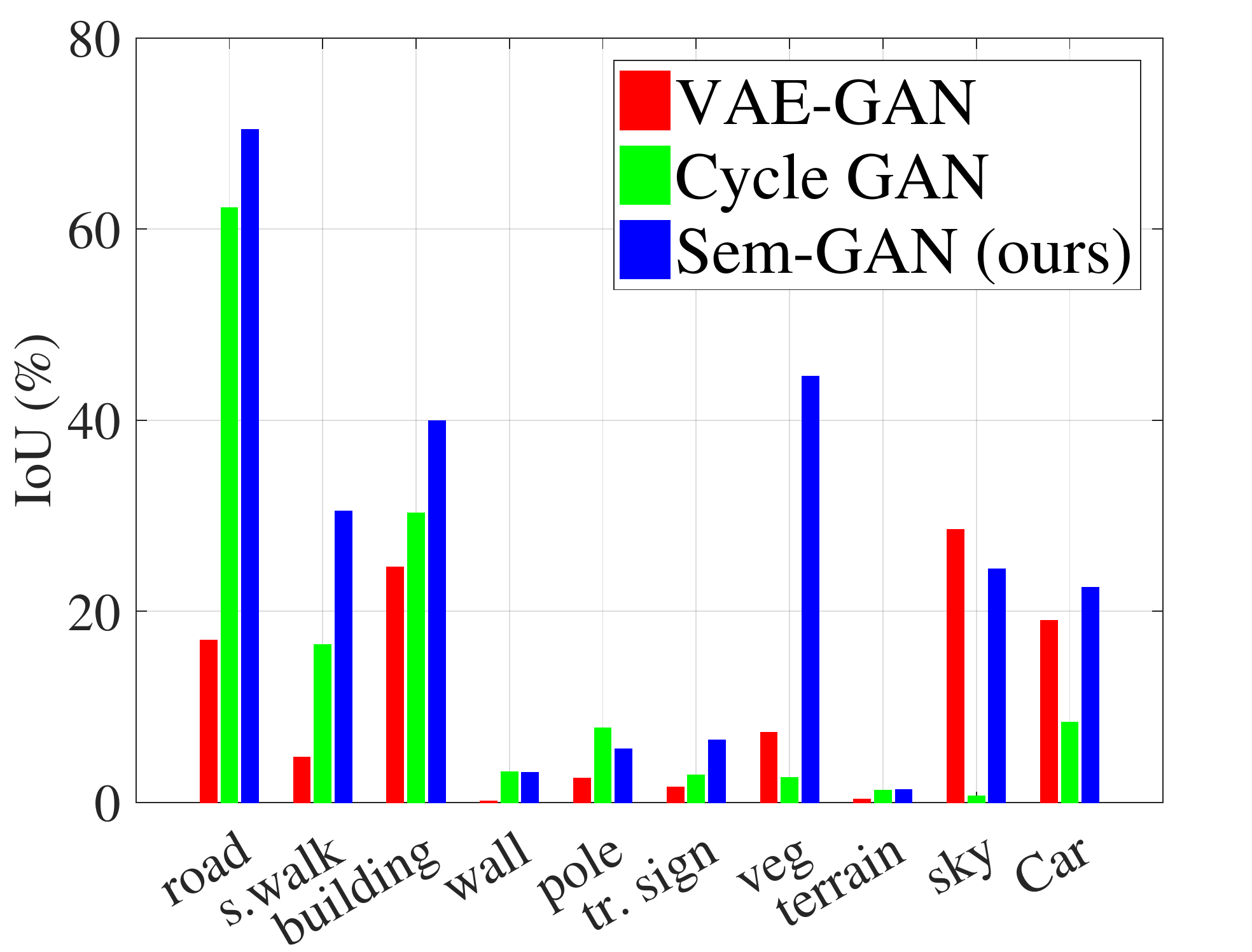}}
	\caption{Per-class IoU scores on the Viper$\leftrightarrow$CS task.}
	\label{fig:viper2cs_barplot}
\end{figure}

\begin{figure}[h]
	\centering
	\subfigure[MP$\to$CS]{\includegraphics[width=7.5cm,trim={0.5cm 7cm 0.5cm 7.5cm},clip]{figs/mms2cs_compare_classes_mms2cs_A2B_suppl.pdf}}
	\subfigure[CS$\to$Viper]{\includegraphics[width=7.5cm,trim={0.5cm 7cm 0.5cm 7.5cm},clip]{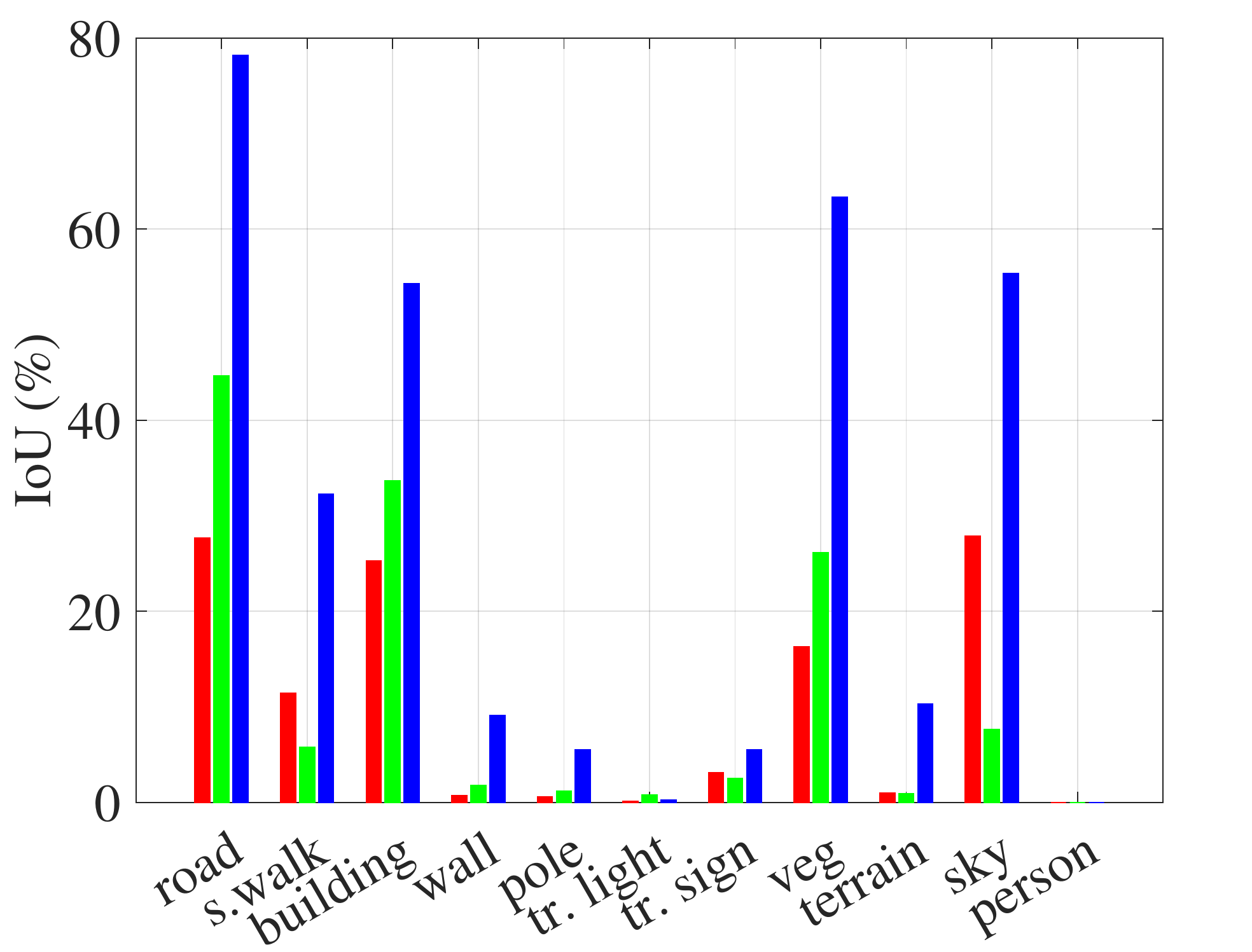}}
	\caption{Per-class IoU scores on the MP$\leftrightarrow$CS task.}
	\label{fig:mms2cs_barplot}
\end{figure}

\begin{figure}[h]
	\centering
	\subfigure[Summer(CS)$\to$Winter (MP)]{\includegraphics[width=7.5cm,trim={0.5cm 7cm 0.5cm 7.5cm},clip]{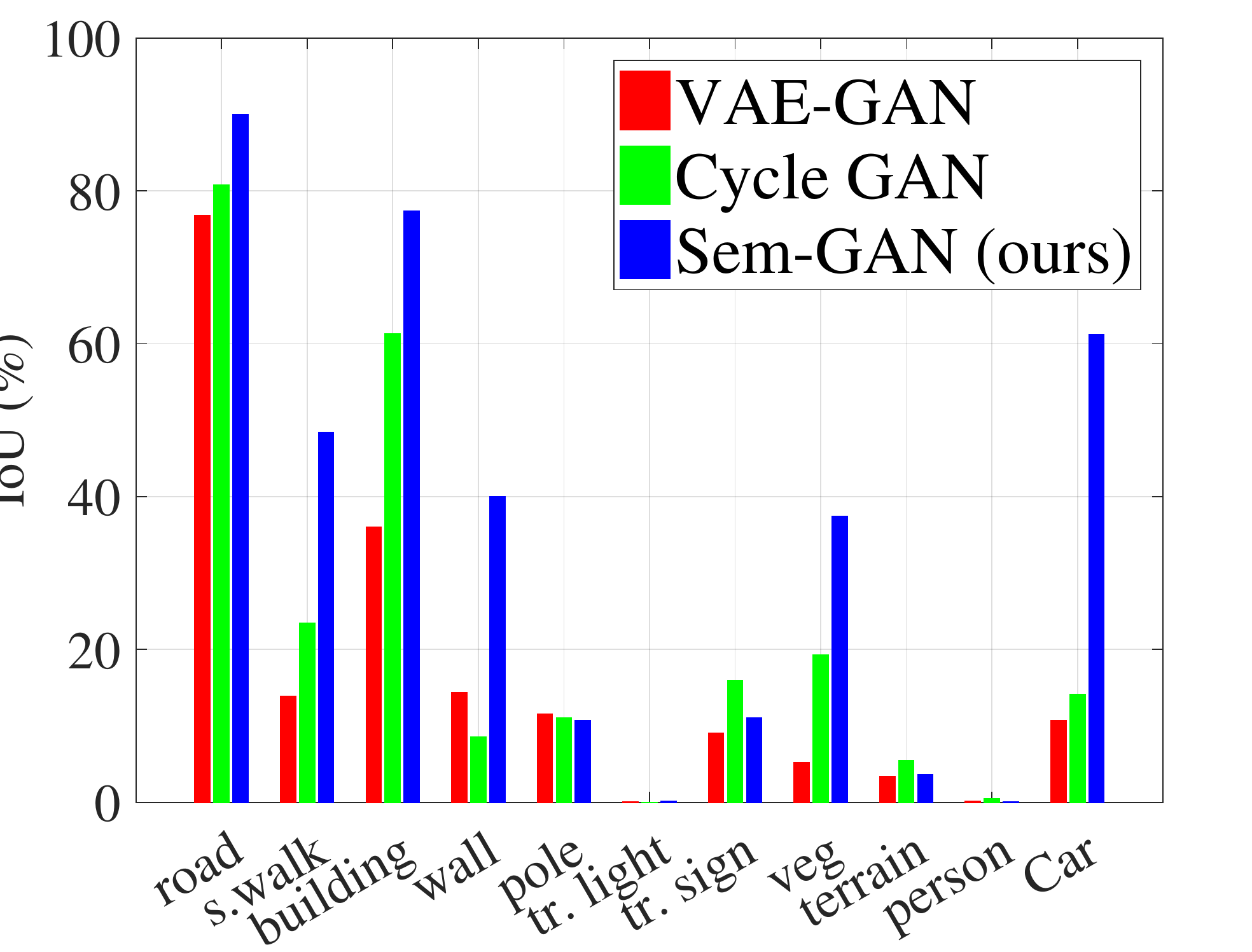}}
	\subfigure[Winter (MP)$\to$Summer (CS)]{\includegraphics[width=7.5cm,trim={0.5cm 7cm 0.5cm 7.5cm},clip]{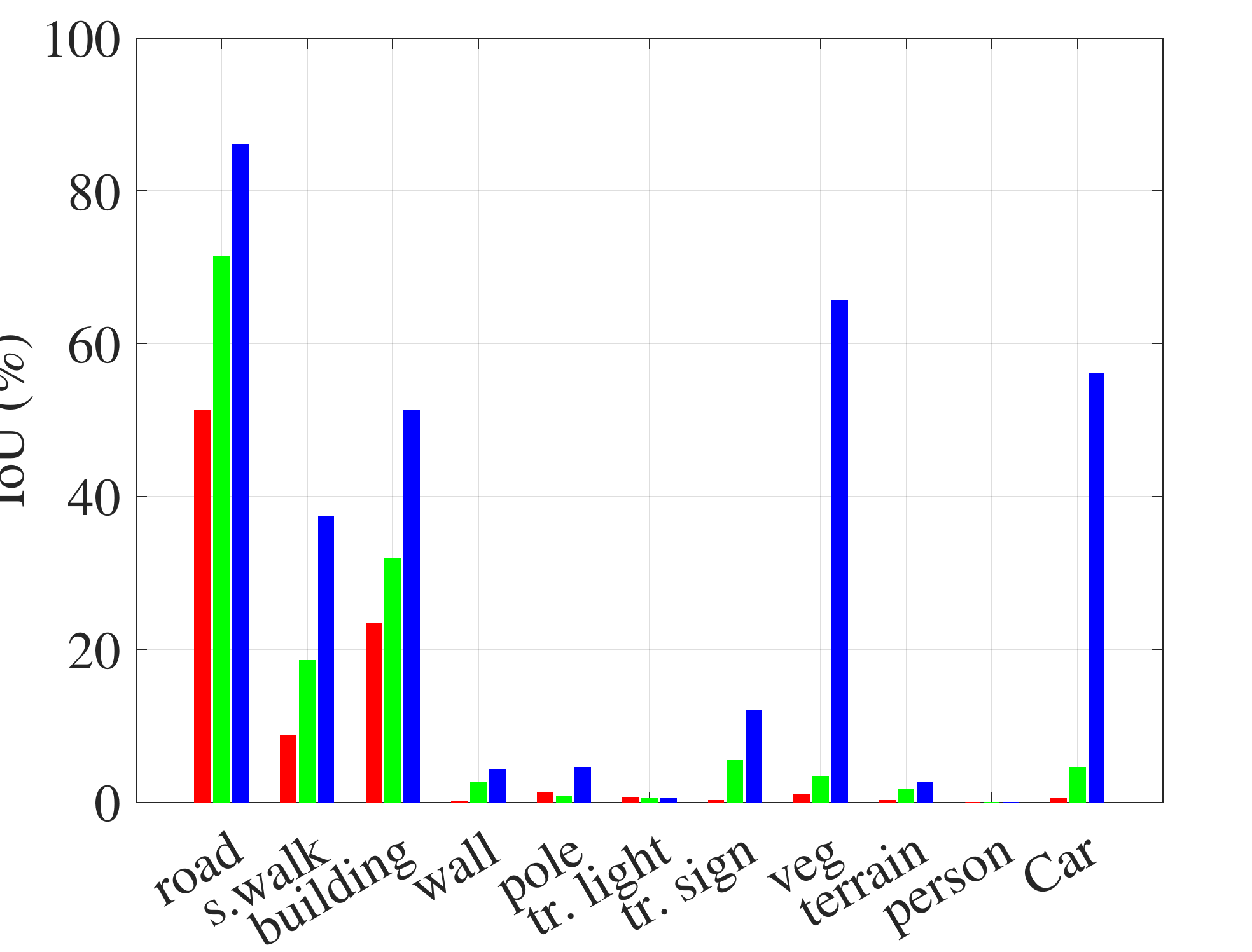}}
	\caption{Per-class IoU scores on the CS summer$\leftrightarrow$MP winter task.}
	\label{fig:cs_sum2win_barplot}
\end{figure}

\begin{figure}[!htbp]
	\centering
	\subfigure[Seg$\to$CS]{\includegraphics[width=7.5cm,trim={0.5cm 7cm 0.5cm 7.5cm},clip]{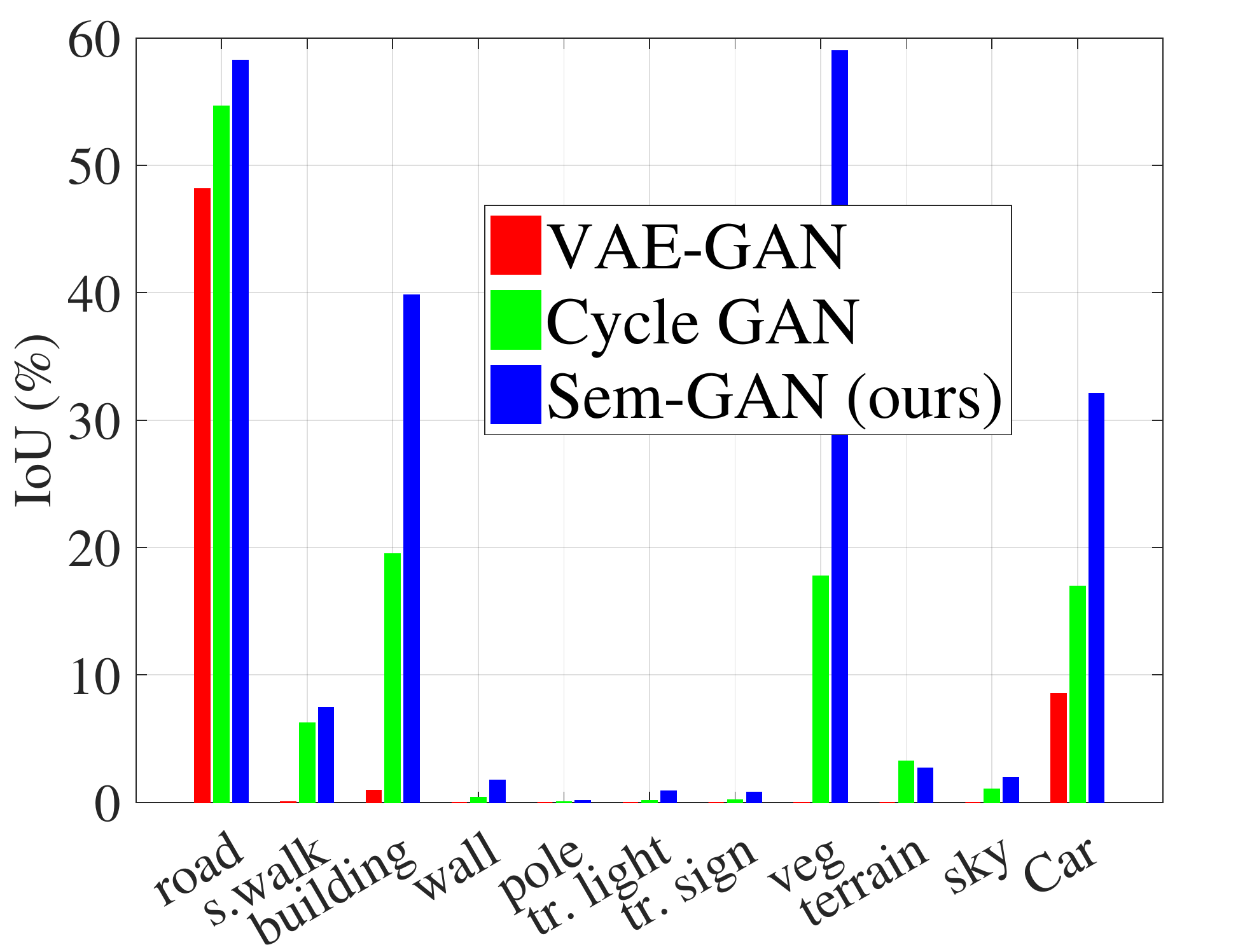}}
	\caption{Per-class IoU scores on the Segmentation mask $\leftrightarrow$ Image (CS) task.}
	\label{fig:seg2im_barplot}
\end{figure}

As mentioned in the main paper, we use 19 semantic segment classes with respect to the Cityscapes dataset for training our Sem-GAN framework. These classes are as follows: 1. 'road', 2.'sidewalk',	3.'building', 4. 'wall', 5. 'fence', 6. 'pole', 7. 'traffic light', 8. 'traffic sign', 9. 'vegetation', 10. 'terrain', 11. 'sky', 12. 'person', 13. 'rider', 14. 'car', 15. 'truck', 16. 'bus', 17. 'train', 18. 'motorcycle', 19. 'bicycle'. Below, we provide the per-class IoU for the following tasks: Viper$\leftrightarrow$CS Figure~\ref{fig:viper2cs_barplot}, MP$\leftrightarrow$CS Figure~\ref{fig:mms2cs_barplot}, CS summer $\leftrightarrow$ MP winter Figure~\ref{fig:cs_sum2win_barplot}, and Seg$\to$ Image (CS) Figure~\ref{fig:seg2im_barplot}.


\section{Ablative Analysis}
In Table~\ref{tab:ablative}, we provide an ablative study of the various elements in our framework. Interestingly, we find that adding the segmentation information into the translation process significantly improves the accuracy from Cycle-GAN, and `no cycle + seg' is about 12\% better than with cycle. This is perhaps because having segmentation information makes the translation process `easier', while without that the cycle-GAN has to figure out the mapping between various segments automatically, which may lead to incorrect mappings. Adding cycle consistency still improves the performance, and seg+cycle+SM performs the best.\footnote{Note that, when we say `no cycle', we mean that we do not use both the cycle-consistency and the identity constraint, as in the implementation of Cycle-GAN.}

\begin{table}[hb]
	\centering
	\begin{adjustbox}{width=1\linewidth}
		\renewcommand{\arraystretch}{1} 
	\begin{tabular}{l|c|c||c|c}
		component & mean Acc. & mIoU & mean Acc.  & mIoU\\
		\hline
		Cycle GAN &	23.6	&	9.2	&	21.0	&	13.4\\
		%No Cycle + With Seg	&  45.6	&	27.5	&	31.9	&	24.3\\
		No Cycle + With Seg	&  35.6	&	22.5	&	25.9	&	19.3\\
		Cycle + Seg &	38.8	&	24.0 &	27.4	&	20.4\\
		Cycle + Seg + SM &	42.5	&	\textbf{28.4} & 	27.7&	\textbf{21.5}\\
	\end{tabular}
	\end{adjustbox}
	\caption{Ablative study of the influence of various components in our model on the Viper2CS task. Left part of the table is the translation from Viper$\to$CS and the right side is CS$\to$Viper. All numbers are in \%.}
	\label{tab:ablative}
\end{table}

\begin{table*}[t]
	\centering
	\begin{adjustbox}{width=1\linewidth}
		\renewcommand{\arraystretch}{1.2} 
		\begin{tabular}{|l|l|l|l|l|l|l|l|l|l|l|l|l|l|l|l|l|l|l|l|l|c|c|c|c}
			\hline
			Method &\rb{Architecture} & \rb{Road} & \rb{sidewalk} & \rb{Building} & \rb{wall} & \rb{fence} & \rb{pole} & \rb{tra. light} & \rb{tra. sign} & \rb{veg.} & \rb{terrain} & \rb{sky} & \rb{person} & \rb{rider} &  \rb{car} & \rb{truck} & \rb{bus} & \rb{train} & \rb{mo. cycle} & \rb{bicycle}& \tb{mIoU} &  {Ov. Acc} & {Fq.W. Acc}\\
			\hline
			CS only & \tb{F} & 85.1 & 38.9 & 60.6 & 0.8 & 1.1 & 0.0 & 0.0 & 0.0 & 65.1 & 7.4 & 36.3 & 19.8 & 0.0 & 62.7 & 0.0 & 0.0 & 0.0 & 0.0 & 0.4 &  19.9 &   77.4 &  64.1 \\
			\hline
			% MP only & F\\
			% SemGAN(MP) & F \\
			CS + MP & F & 87.0 & 41.8 & 64.6 & 14.5 & 0.2 & 0.6 & 0.1 & 0.4 & 68.6 & 11.7 & 67.3 & 11.2 & 0.0 & 63.6 & 0.0 & 0.0 & 0.0 & 0.0 & 0.1 &  22.7 &    79.3 &  66.8 \\
			CS+Cy(MP) & F & 85.5 & 40.3 & 63.6 & 6.9 & 0.0 & 3.2 & 0.4 & 3.5 & 69.0 & 7.8 & 52.2 & 11.7 & 0.0 & 62.5 & 0.0 & 0.0 & 0.0 & 0.0 & 2.6 &  21.5 &   77.8 &  65.6 \\
			CS+Sm(MP) & F & 88.1 & 47.0 & 67.8 & 12.8 & 0.5 & 7.1 & 0.0 & 2.0 & 71.1 & 10.0 & 69.0 & 15.4 & 0.0 & 67.6 & 0.0 & 0.0 & 0.0 & 0.0 & 4.0 &  \tb{24.3} &   80.8 &  69.1 \\
			\hline
			% VP only & F 
			% SemGAN(VP) & F \\
			CS + VP & F & 90.2 & 54.1 & 70.3 & 22.6 & 5.9 & 3.4 & 0.0 & 3.4 & 73.4 & 27.2 & 67.1 & 31.4 & 0.3 & 73.2 & 9.1 & 18.7 & 5.6 & 0.2 & 21.2 &  30.4 &    83.2 &  72.6 \\ 
			CS+Cy(VP) & F & 90.3 & 53.2 & 67.4 & 9.4 & 15.3 & 2.6 & 0.1 & 4.8 & 70.6 & 26.4 & 60.9 & 36.4 & 0.9 & 74.2 & 21.1 & 24.9 & 3.4 & 5.0 & 30.4 &  31.4  &  82.8 &  71.9\\
			CS+Sm(VP) &F & 92.1 & 59.1 & 71.3 & 21.6 & 19.1 & 4.4 & 0.2 & 5.6 & 74.1 & 30.2 & 70.1 & 36.4 & 1.3 & 76.8 & 24.2 & 20.5 & 11.7 & 4.3 & 30.7 &  \tb{34.4} &  84.6 &  74.9\\
			\hline
			\hline
			CS only & \tb{P} & 85.2 & 35.2 & 62.9 & 4.1 & 15.4 & 0.5 & 0.0 & 2.6 & 68.6 & 24.3 & 49.0 & 27.2 & 0.0 & 63.6 & 0.0 & 4.8 & 0.0 & 0.0 & 19.7 &  24.4 &  78.7 &  65.7\\
			\hline
			MP only & P & 51.2 & 12.9 & 40.7 & 4.6 & 0.0 & 4.8 & 0.2 & 9.3 & 50.5 & 2.6 & 10.3 & 0.0 & 0.0 & 59.5 & 0.0 & 0.0 & 0.0 & 0.0 & 0.0 &  13.0 &  56.0 &  42.0 \\
			Cy(MP) & P & 83.2 & 32.2 & 49.3 & 7.3 & 0.0 & 5.1 & 0.8 & 14.0 & 43.9 & 10.0 & 28.1 & 0.0 & 0.0 & 50.5 & 0.0 & 0.0 & 0.0 & 0.0 & 0.0 &  17.1 &    69.1 &  56.4 \\
			Sm(MP) & P & 85.4 & 40.9 & 63.8 & 11.9 & 0.0 & 8.5 & 1.2 & 10.0 & 69.9 & 13.9 & 64.0 & 0.0 & 0.0 & 63.2 & 0.0 & 0.0 & 0.0 & 0.0 & 0.0 &  \tb{22.8} &    78.2 &  66.0\\
			\hline
			CS+MP & P & 88.8 & 49.5 & 70.2 & 7.0 & 4.7 & 9.0 & 0.0 & 13.4 & 72.8 & 20.2 & 74.9 & 38.2 & 0.0 & 73.5 & 0.0 & 3.2 & 0.0 & 0.0 & 31.5 &  29.3 &    82.4 &  71.7\\
			CS+Cy(MP) & P & 89.8 & 51.3 & 71.3 & 14.1 & 4.2 & 11.2 & 0.7 & 17.9 & 73.3 & 23.7 & 63.5 & 39.2 & 0.7 & 73.2 & 0.0 & 7.6 & 0.0 & 0.0 & 34.7 &  30.3 &   72.6 &  83.3\\
			CS+Sm(MP) &P & 90.3 & 54.2 & 72.4 & 17.4 & 8.0 & 16.6 & 0.1 & 17.9 & 75.8 & 23.6 & 74.2 & 42.8 & 8.5 & 74.3 & 0.0 & 17.3 & 0.0 & 0.0 & 36.1 &  \tb{33.1} &    84.3 &  74.1\\
			\hline
			VP only & P & 75.9 & 29.4 & 49.4 & 0.0 & 0.0 & 0.0 & 0.6 & 10.8 & 65.2 & 13.2 & 62.5 & 15.7 & 0.0 & 58.3 & 12.7 & 6.2 & 0.0 & 0.1 & 0.0 & 21.1 & 71.8 & 57.8\\
			Cy(VP) & P & 85.2 & 35.2 & 53.6 & 0.0 & 4.7 & 0.0 & 4.3 & 9.6 & 22.3 & 15.3 & 20.3 & 11.3 & 0.0 & 66.4 & 11.8 & 3.0 & 0.0 & 4.1 & 0.0 &  18.3 &    70.9 &  56.8 \\
			Sm(VP) & P & 87.9 & 37.0 & 56.5 & 0.0 & 4.5 & 0.0 & 2.2 & 15.6 & 42.6 & 24.8 & 40.4 & 20.0 & 0.0 & 74.1 & 19.9 & 14.7 & 0.0 & 4.6 & 0.0 &  \tb{23.4} &    76.2 &  62.8 \\
			\hline
			CS+VP & P & 91.5 & 54.2 & 74.8 & 23.8 & 7.4 & 18.3 & 3.0 & 13.7 & 76.9 & 24.8 & 66.5 & 48.6 & 22.2 & 82.1 & 35.7 & 19.7 & 28.2 & 6.4 & 42.7 &  39.0 &    85.6 &  76.3 \\
			CS+Sm(VP) & P & 93.4 & 63.4 & 76.4 & 27.3 & 11.7 & 23.6 & 15.8 & 23.6 & 78.2 & 32.0 & 78.4 & 52.2 & 26.2 & 84.0 & 33.4 & 33.4 & 30.1 & 18.1 & 42.4 &  \tb{44.4} &    87.4 &  79.1\\ 
			% CS+CyGAN(VP) & P\\
			\hline
		\end{tabular}
	\end{adjustbox}
	\caption{Training segmentation models using adapted images. We adapt synthetic MP and VIPER (VP) datasets to Cityscapes (CS) domain. Sm(MP) and Cy(MP) denote adaptation of all images in MP with SemGAN and Cycle GAN respectively to the CS domain. "only" refers to using images directly from that domain (without adaptation). We use two segmentation CNNs: "F" is VGG-FCN8s and "P" is PSPNet.}
	\label{tab:seg_acc}
\end{table*}

\section{When Ground Truth Masks are Unavailable}
\begin{figure*}
	\centering
	\subfigure[Horse$\to$Zebra]{\includegraphics[width=4.8cm]{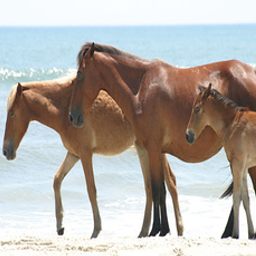}}
	\subfigure[Cycle GAN]{\includegraphics[width=4.8cm]{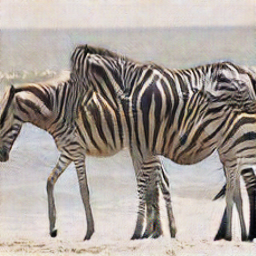}}
	\subfigure[Sem GAN]{\includegraphics[width=4.8cm]{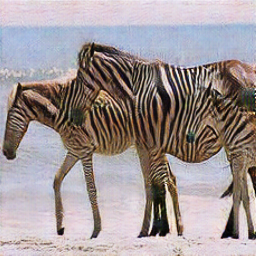}}\\
	\subfigure[Zebra$\to$Horse]{\includegraphics[width=4.8cm]{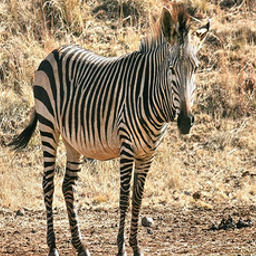}}
	\subfigure[Cycle GAN]{\includegraphics[width=4.8cm]{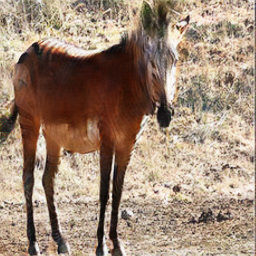}}
	\subfigure[Sem GAN]{\includegraphics[width=4.8cm]{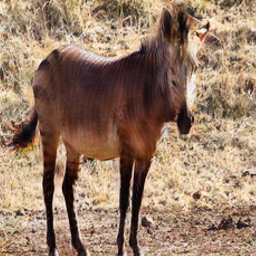}}\\
	\subfigure[Horse$\to$Zebra]{\includegraphics[width=4.8cm]{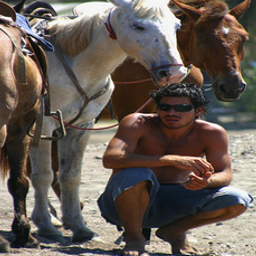}}
	\subfigure[Cycle GAN]{\includegraphics[width=4.8cm]{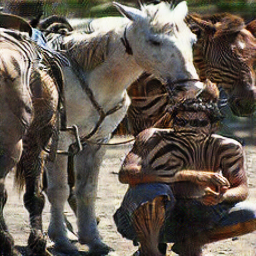}}
	\subfigure[Sem GAN]{\includegraphics[width=4.8cm]{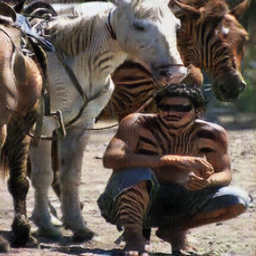}}\\	
	\subfigure[Horse$\to$Zebra]{\includegraphics[width=4.8cm]{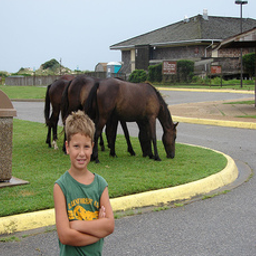}}
	\subfigure[Cycle GAN]{\includegraphics[width=4.8cm]{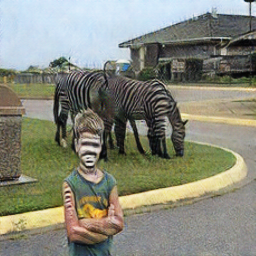}}
	\subfigure[Sem GAN]{\includegraphics[width=4.8cm]{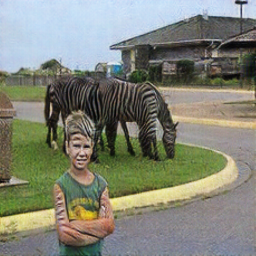}}
	\caption{Translation from `horse' to `zebra'. Here we use segmentation model trained on MS-COCO dataset as the images in the task does not come with their semantic labels.}
	\label{fig:horse2zebra}
\end{figure*}

\vspace*{-0.2cm}
As alluded to in the main paper, we do not necessarily require the ground truth semantic masks for our scheme to work -- instead we only need to have a segmentation model for the respective domains. To this end, we experiment this facet of our scheme on the task of translating 'horses' to 'zebras' using the dataset provided with Cycle-GAN~\cite{zhu2017unpaired}. There are about 1300 images of horses and zebras in this dataset. For the segmentation models we use an FCN network trained on the MSCOCO dataset that has 80 semantic classes including `horse' and 'zebra'. We do not train these models within our Sem-GAN setup. Qualitative results from this experiment are provided in Figure~\ref{fig:horse2zebra}. To ensure the translations are cross-domain, that is the source is, say the 'horse' class and the target is the 'zebra' class, for defining the consistency criteria, we \textbf{switch the labels} of the source segmenter (which in this case will identify 'horse') to 'zebra', and vice versa for the other translation direction. For this task, we trained both Cycle GAN and Sem GAN for 200 epochs. We used a 9-block ResNet for the generator.

A point to be noted in this task is that, while the results of both Cycle-GAN and Sem-GAN are more or less similar, the translation with Sem-GAN is slightly better (qualitatively) when multiple classes are present in the images -- such humans (see for example, the last two rows in Figure~\ref{fig:horse2zebra}). This is because, the MS-COCO segmentation dataset includes a `person' class as well. While, the results seem better, there still remains a lot to improve; especially to get the structure of the objects within a segment.

\section{Additional Results on Semantic Segmentation Task}
In addition to the results in Table 3 in the main paper, in Table~\ref{tab:seg_acc} we provide additional results on the semantic segmentation task using synthetic images (translated using Cycle-GAN or Sem-GAN)) for training segmentation models. The additional results are for segmentation models trained only on translated synthetic images (not using real images from the domain or their ground truths) -- such as Cy(VP) and Sm(VP). Interestingly, we find that using only Sm(VP) is better than using VP only (21.1 against 23.4\%) and using MP only to Sm(MP) is increased from 13.0 to 22.8\% in mIoU clearly demonstrating that our Sem-GAN leads to much better domain adaptations than using the synthetic images directly. We also see that Cy(MP) and Cy(VP) are inferior in performance.

\section{Qualitative Results}
From Figure~\ref{fig:mms2cs_quals} onwards, we provide additional qualitative results on the tasks we described in the main paper.

%
%\begin{figure*}
%	\centering
%	\subfigure[Real(CS)$\to$ Syn.(MP)]{\includegraphics[width=4.8cm]{suppl_quals/mms2cs/2_4_real_B.png}}
%	\subfigure[Cycle GAN]{\includegraphics[width=4.8cm]{suppl_quals/mms2cs/2_4_fake_A_orig.png}}
%	\subfigure[Sem GAN]{\includegraphics[width=4.8cm]{suppl_quals/mms2cs/2_4_fake_A_seg.png}}\\	
%	\subfigure[Real(CS)$\to$ Syn.(MP)]{\includegraphics[width=4.8cm]{suppl_quals/mms2cs/2_23_real_B.png}}
%	\subfigure[Cycle GAN]{\includegraphics[width=4.8cm]{suppl_quals/mms2cs/2_23_fake_A_orig.png}}
%	\subfigure[Sem GAN]{\includegraphics[width=4.8cm]{suppl_quals/mms2cs/2_23_fake_A_seg.png}}\\
%	\subfigure[Real(CS)$\to$ Syn.(MP)]{\includegraphics[width=4.8cm]{suppl_quals/mms2cs/2_70_real_B.png}}
%	\subfigure[Cycle GAN]{\includegraphics[width=4.8cm]{suppl_quals/mms2cs/2_70_fake_A_orig.png}}
%	\subfigure[Sem GAN]{\includegraphics[width=4.8cm]{suppl_quals/mms2cs/2_70_fake_A_seg.png}}\\
%	\subfigure[Real(CS)$\to$ Syn.(MP)]{\includegraphics[width=4.8cm]{suppl_quals/mms2cs/2_89_real_B.png}}
%	\subfigure[Cycle GAN]{\includegraphics[width=4.8cm]{suppl_quals/mms2cs/2_89_fake_A_orig.png}}
%	\subfigure[Sem GAN]{\includegraphics[width=4.8cm]{suppl_quals/mms2cs/2_89_fake_A_seg.png}}\\
%	\caption{Translation from Real (Cityscapes) to Synthetic (MP). We show the real image (left), the translation by Cycle GAN (middle) and that by Sem GAN (right).}
%	\label{fig:mms2cs_quals}
%\end{figure*}

\begin{figure*}
	\centering
	\subfigure[Syn(Viper)$\to$ CS]{\includegraphics[width=4.8cm]{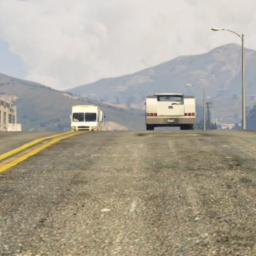}}
	\subfigure[Cycle GAN]{\includegraphics[width=4.8cm]{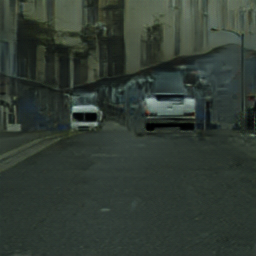}}
	\subfigure[Sem GAN]{\includegraphics[width=4.8cm]{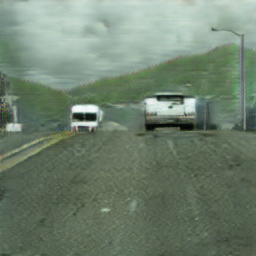}}\\
	\subfigure[Syn(Viper)$\to$ CS]{\includegraphics[width=4.8cm]{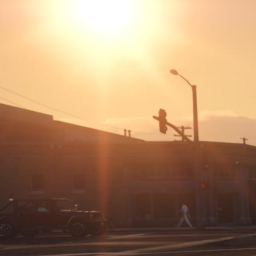}}
	\subfigure[Cycle GAN]{\includegraphics[width=4.8cm]{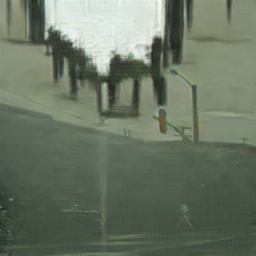}}
	\subfigure[Sem GAN]{\includegraphics[width=4.8cm]{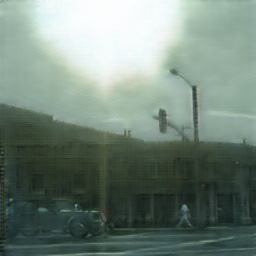}}\\
	\subfigure[Syn(Viper)$\to$ CS]{\includegraphics[width=4.8cm]{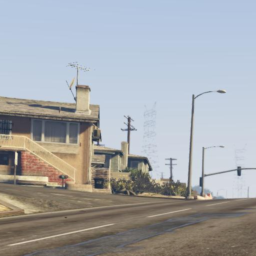}}
	\subfigure[Cycle GAN]{\includegraphics[width=4.8cm]{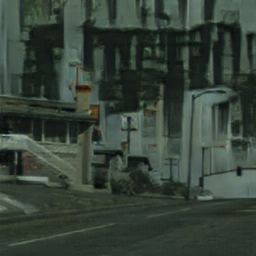}}
	\subfigure[Sem GAN]{\includegraphics[width=4.8cm]{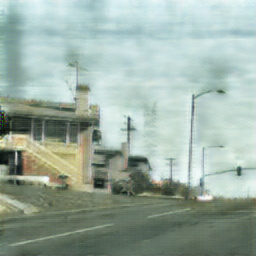}}\\	
	\subfigure[Syn(Viper)$\to$ CS]{\includegraphics[width=4.8cm]{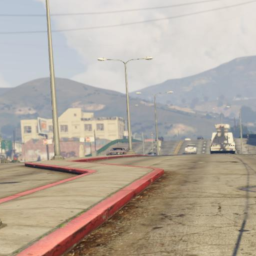}}
	\subfigure[Cycle GAN]{\includegraphics[width=4.8cm]{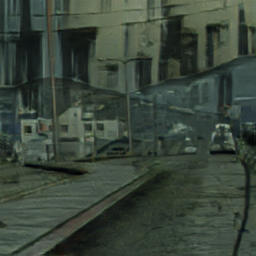}}
	\subfigure[Sem GAN]{\includegraphics[width=4.8cm]{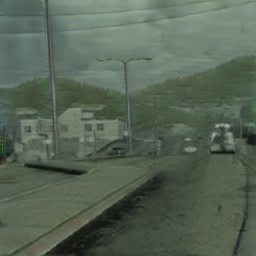}}\\
	\caption{Translation from  Synthetic (Viper) to Real (CS). We show the viper image (left), the translation by Cycle GAN (middle) and that by Sem GAN (right).}
\end{figure*}

\begin{figure*}
	\centering
	\subfigure[Seg$\to$image (CS)]{\includegraphics[width=4.8cm]{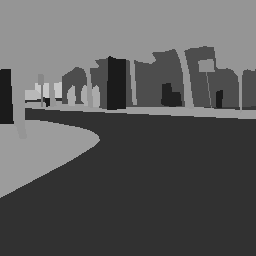}}
	\subfigure[Cycle GAN]{\includegraphics[width=4.8cm]{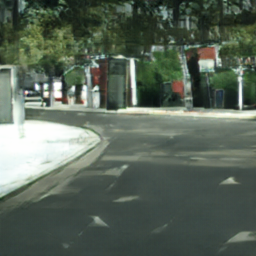}}
	\subfigure[Sem GAN]{\includegraphics[width=4.8cm]{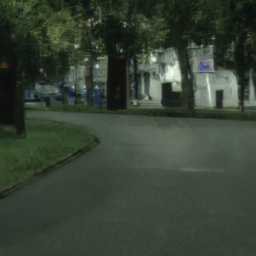}}\\
	\subfigure[Seg$\to$image (CS)]{\includegraphics[width=4.8cm]{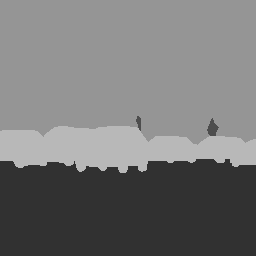}}
	\subfigure[Cycle GAN]{\includegraphics[width=4.8cm]{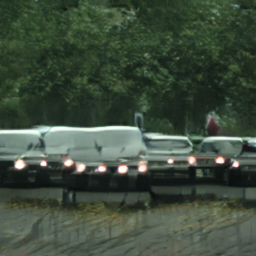}}
	\subfigure[Sem GAN]{\includegraphics[width=4.8cm]{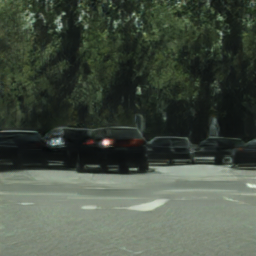}}\\
	\subfigure[Seg$\to$image (CS)]{\includegraphics[width=4.8cm]{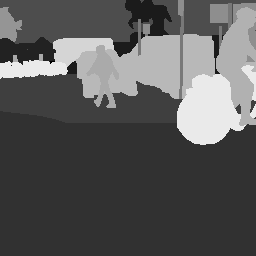}}
	\subfigure[Cycle GAN]{\includegraphics[width=4.8cm]{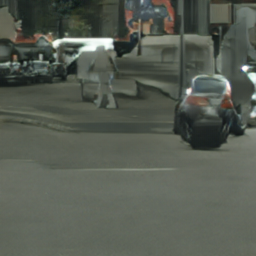}}
	\subfigure[Sem GAN]{\includegraphics[width=4.8cm]{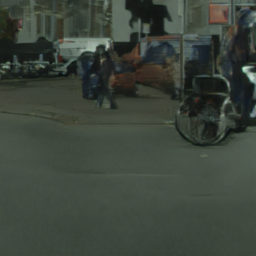}}\\
	\subfigure[Seg$\to$image (CS)]{\includegraphics[width=4.8cm]{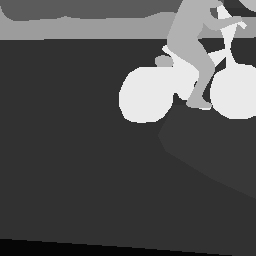}}
	\subfigure[Cycle GAN]{\includegraphics[width=4.8cm]{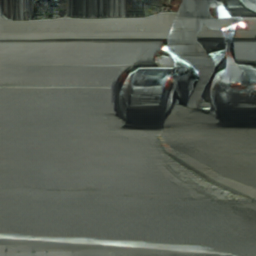}}
	\subfigure[Sem GAN]{\includegraphics[width=4.8cm]{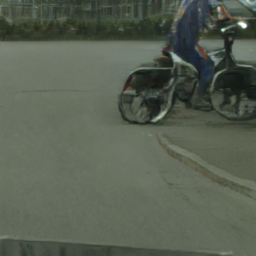}}\\
	\caption{Translation from segmentation mask to Real (cityscapes). We show the segmask (left), the translation by Cycle GAN (middle) and that by Sem GAN (right).}
\end{figure*}

\begin{figure*}
	\centering
	\subfigure[CS(day)$\to$ night]{\includegraphics[width=4.8cm]{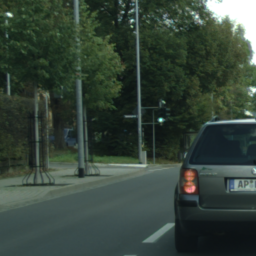}}
	\subfigure[Sem GAN]{\includegraphics[width=4.8cm]{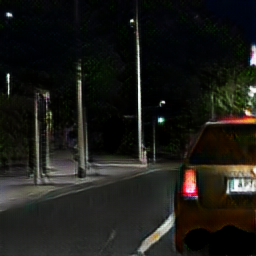}}\\
	\subfigure[CS(day)$\to$ night]{\includegraphics[width=4.8cm]{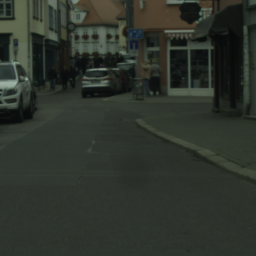}}
	\subfigure[Sem GAN]{\includegraphics[width=4.8cm]{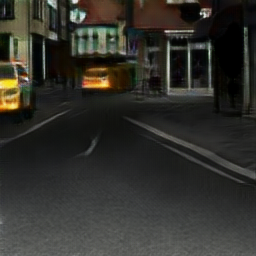}}\\
	\subfigure[CS(day)$\to$ night]{\includegraphics[width=4.8cm]{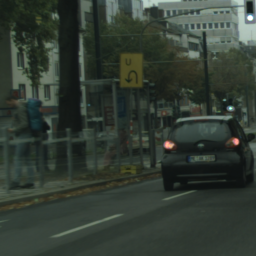}}
	\subfigure[Sem GAN]{\includegraphics[width=4.8cm]{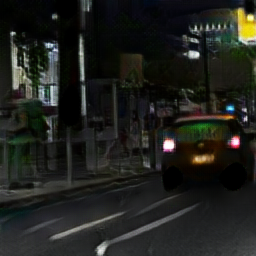}}\\
	\subfigure[CS(day)$\to$ night]{\includegraphics[width=4.8cm]{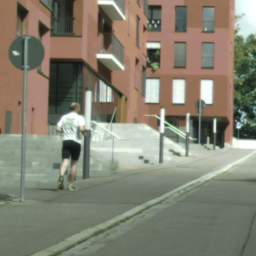}}
	\subfigure[Sem GAN]{\includegraphics[width=4.8cm]{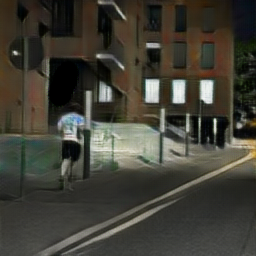}}\\
	\caption{Translation from cityscapes real image (left) to MP synthetic night image (right).}
\end{figure*}
\begin{figure*}
	\centering
	\subfigure[Syn(night)$\to$ CS(day)]{\includegraphics[width=4.8cm]{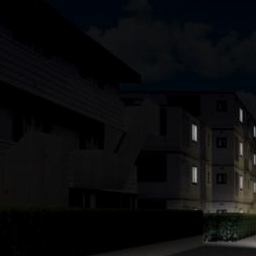}}
	\subfigure[Sem GAN]{\includegraphics[width=4.8cm]{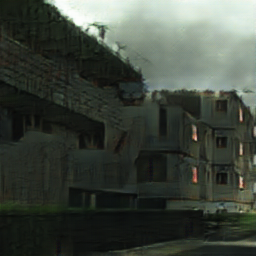}}\\
	\subfigure[Syn(night)$\to$ CS(day)]{\includegraphics[width=4.8cm]{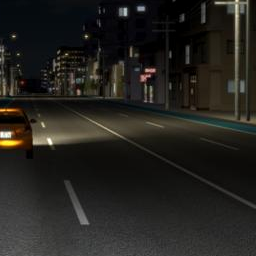}}
	\subfigure[Sem GAN]{\includegraphics[width=4.8cm]{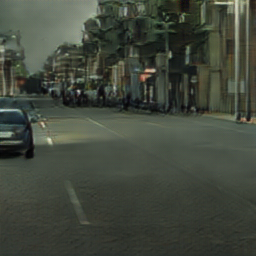}}\\
	\caption{Translation from MP synthetic night image (left) to cityscapes real image (right).}
\end{figure*}
\begin{figure*}
	\centering
	\subfigure[Real(CS)$\to$Winter]{\includegraphics[width=4.8cm]{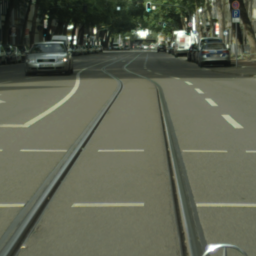}}
	\subfigure[Synthetic Winter]{\includegraphics[width=4.8cm]{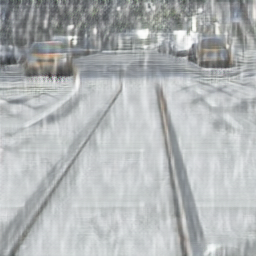}}\\
	\subfigure[Real(CS)$\to$Winter]{\includegraphics[width=4.8cm]{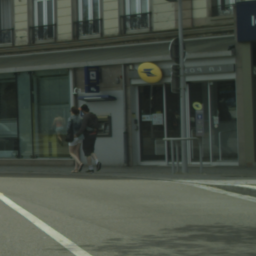}}
	\subfigure[Synthetic Winter]{\includegraphics[width=4.8cm]{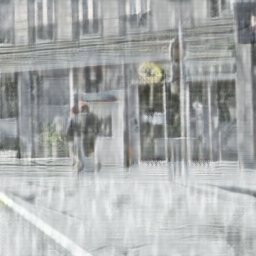}}\\
	\subfigure[Winter(MP)$\to$CS]{\includegraphics[width=4.8cm]{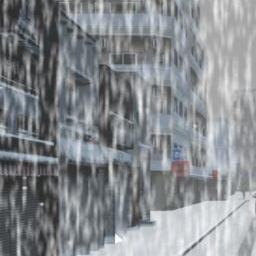}}
	\subfigure[Sem GAN]{\includegraphics[width=4.8cm]{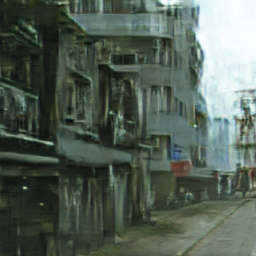}}\\
	\subfigure[Winter(MP)$\to$CS]{\includegraphics[width=4.8cm]{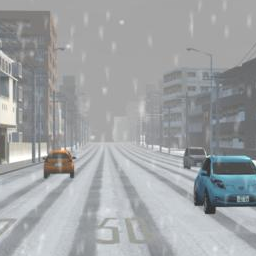}}
	\subfigure[Sem GAN]{\includegraphics[width=4.8cm]{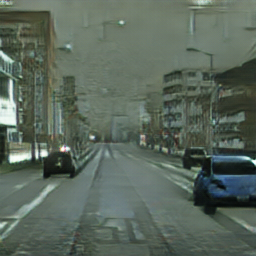}}
	\caption{Rows 1--2: Translation from cityscapes real image (left) to MP synthetic winter image (right). Rows 3--4: Translation from MP synthetic winter image (left) to real (CS) domain.}
\end{figure*}

\end{document}